\RequirePackage{snapshot}
\documentclass[preprint,11pt,authoryear]{elsarticle}
\usepackage{amsmath}
\usepackage{amsfonts}
\usepackage{amssymb}
\usepackage{mathrsfs}
\usepackage{graphicx}
\usepackage[margin=1in]{geometry}
\usepackage[hidelinks]{hyperref}
\usepackage[american]{babel}
\usepackage{multirow}
\usepackage{pifont}
\usepackage{float}
\usepackage{natbib}
\usepackage{tipa}
\usepackage{tikz}

\begin{document}

\begin{frontmatter}

\title{CiwGAN and fiwGAN: Encoding information in acoustic data to model lexical learning with Generative Adversarial Networks}
\author[1]{Ga\v{s}per Begu\v{s}}
\ead{begus@berkeley.edu}

  \address[1]{Department of Linguistics, University of California, Berkeley, 1203 Dwinelle Hall \#2650, Berkeley, CA 94720 }

\journal{Neural Networks}

\begin{abstract}

How can deep neural networks encode information that corresponds to words in human speech into raw acoustic data? This paper proposes two neural network architectures for modeling unsupervised lexical learning from raw acoustic inputs: ciwGAN (Categorical InfoWaveGAN) and fiwGAN (Featural InfoWaveGAN). These combine Deep Convolutional GAN architecture for audio data (WaveGAN; \citealt{donahue19}) with the information theoretic extension of GAN --- InfoGAN \citep{chen16} --- and propose a new latent space structure that can model featural learning simultaneously with a higher level classification and allows for a very low-dimension vector representation of lexical items. In addition to the Generator and Discriminator networks, the architectures introduce a network that learns to retrieve latent codes from generated audio outputs. Lexical learning is thus modeled as emergent from an architecture that forces a deep neural network to output data such that unique information is retrievable from its acoustic outputs. The networks trained on lexical items from the TIMIT corpus learn to encode unique information corresponding to lexical items in the form of categorical variables in their latent space. By manipulating these variables, the network outputs specific lexical items. The network occasionally outputs innovative lexical items that violate training data, but are linguistically interpretable and highly informative for cognitive modeling and neural network interpretability. Innovative outputs suggest that phonetic and phonological representations learned by the network can be productively recombined and directly paralleled to productivity in human speech: a fiwGAN network trained on \emph{suit} and \emph{dark} outputs innovative \emph{start}, even though it never saw \emph{start} or even a [st] sequence in the training data. We also argue that setting latent featural codes to values well beyond training range results in almost categorical generation of prototypical lexical items and reveals underlying values of each latent code. Probing deep neural networks trained on well understood dependencies in speech bears implications for latent space interpretability and  understanding how deep neural networks learn meaningful representations, as well as potential for unsupervised text-to-speech generation in the GAN framework. 
\end{abstract}




\end{frontmatter}
\section{\label{intro} Introduction}

How human language learners encode information in their speech is among the core questions in linguistics and computational cognitive science. Acoustic speech data is the primary  source of linguistic input for hearing infants, and first language learners must learn to retrieve information from raw acoustic data. By the time language acquisition is complete, learners are able to not only analyze but also produce speech consisting of words (henceforth lexical items) that carry meaning  \citep{saffran96,saffran06,kuhl10}. In other words, speakers learn to encode information in their acoustic output, and they do so by associating meaning-bearing units of speech (lexical items) with unique information. Lexical items in turn consist of units called \emph{phonemes} that represent individual sounds. In fact, speakers not only produce lexical items that exist in their primary linguistic data, but are also able to generate new lexical items that consist of novel combinations of phonemes that conform to the phonotactic rules of their language. This points to one of the core properties of language: productivity \citep{hockett59,piantadosi17,baroni19}. 

\subsection{Prior work}
\label{prior}

Computational approaches to lexical learning have a long history. Modeling lexical learning can take many forms (for a comprehensive overview, see \citealt{rasanen12}), but the shift towards modeling lexical learning from acoustic data, especially from raw unreduced acoustic data, has occurred relatively recently (\citealt{levin13,lee15,chung16,bajestan18,kamper19,chorowski19,baayen19}, i.a.). Previously, the majority of models operated on either fully abstracted or already simplified features extracted from raw acoustic data. A variety of models have been proposed for this task including, among others, Bayesian and connectionist approaches (see, among others \citealt{goldwater09,feldman09,rasanen12,heymann13,lee12,elsner13,feldman13,lee15,arnold17,kamper17,bajestan18,baayen19,chuang20}).

 As summarized in \cite{lee15}, existing models of lexical learning that take some form of acoustic data as input can be divided into ``spoken term discovery'' models and ``models of word segmentation'' \cite[390]{lee15}. Proposals of the first approach most commonly involve the clustering of similarities in acoustic data to establish a set of phonetic units from which lexical items are then established, again based on clustering. The \emph{word segmentation} models, on the other hand, ``start from unsegmented strings of symbols and attempt to identify subsequences corresponding to lexical items'' (\citealt[390]{lee15};  for evaluation of the models, see also \citealt{levin13}). The models can take as inputs acoustic data pre-segmented at the word level (as in the current paper and \citealt{kamper14,chung16}) or acoustic inputs of unsegmented speech  (e.g.~in \citealt{lee15,rasanen15,rasanen20}).

Weakly supervised and unsupervised  deep neural network models operating on acoustic data have recently been used to model phonetic learning \citep{rasanen16,alishahi17,eloff19,shain19,chung20}, but not learning of phonological processes. 
Evidence for phonemic representation in deep neural networks, for example,  emerges in a weakly supervised model that combines visual and auditory information \citep{alishahi17}. Several prominent autoencoder models that are trained to  represent data in a lower-dimensionality space have recently been proposed  \citep{rasanen16,eloff19,shain19,chorowski19}. Clustering analyses of the reduced space in these autoencoder models suggest that the networks learn approximates to phonetic features. The disadvantage of the autoencoder architecture is that outputs reproduce inputs as closely as possible: the network's outputs are directly connected to its inputs, which is not an ideal setting for language acquisition. Furthermore, current proposals in the autoencoder framework do not model phonological processes, and there is only an indirect relationship between phonetic properties and latent space.

A prominent framework for modeling lexical learning is acoustic word embedding models that include various (mostly unsupervised) methods \citep{levin13,kamper14} including deep neural networks \citep{chung16,hu20,baevski20,niekerk20}. Similar to the phone-level autoencoder models, the goal of the acoustic word embedding models is a parsimonious encoding of lexical items  that maps acoustic input into a fixed dimensionality vector \citep{levin13,chung16}. The models can be used for unsupervised lexical learning and spoken term discovery in low resource languages (e.g.~the Zerospeech challenge; \citealt{dunbar17,dunbar19,dunbar20}). Several models within this framework employ the autoencoder architecture, where the latent space reduced in dimensionality  can serve as a vector representing acoustic lexical items \citep{chung16,kamper19,niekerk20}. \cite{chung16} show that averaged latent representations can correspond to phonetic representations, but in order to get these results they additionally perform dimensionality reduction on the latent space. Similarly, \cite{chorowski19} use a vector quantized variational autoencoder (VQ-VAE) in which the encoder outputs categorical values constituting token identity. \cite{chorowski19} argue that token identities match phoneme identities with relative high frequencies (see also \citealt{niekerk20,chung20}). \cite{chen20} argue that a convolutional encoder outputs a higher quality of audio compared to the RNN architectures that most of the mentioned proposals use.

One of the major contributions of these models is the facilitation of unsupervised automatic speech recognition (ASR) for zero resource languages, which is why their evaluation focuses on metrics such as the word discrimination/error tasks (e.g.~\citealt{levin13,chung16,kamper19,chen20,niekerk20,baevski20}) or naturalness of the outputs (e.g.~\citealt{eloff19,chen20,niekerk20}). A subset of proposals  explores interpretability of the latent space and generated outputs (see \citealt{chung16,chorowski19}), but they focus on the entire latent space rather than on individual variables or their direct influence on generated outputs. Moreover, the acoustic word embedding models still operate with relatively high dimensional vectors (substantially higher than in the fiwGAN architecture; see Section \ref{fiwGANonTimit}) and their interpretation often includes the entire latent vector or requires additional dimensionality reduction techniques. To my knowledge, exploration of how \emph{individual} variables in these vectors correspond to linguistically meaningful units or how we can elicit  categorical behavior (see Section \ref{up}) is absent. 

Finally, while autoencoders are generative and unsupervised, they crucially differ from GANs in that the encoder does have direct access to the data. Additionally, autoencoders  are trained on replicating data rather than on learning to generate data from noise in an unsupervised manner. In other words, autoencoders are unsupervised in terms of encoding data representations in the latent space, but the data generation part (decoders) is supervised in the autoencoder architecture. GANs, on the other hand, are unsupervised also in the sense of data generation.   This distinction is primarily relevant for the cognitive modeling aspect of the proposal.

The ideal cognitive model of lexical learning would include both the production and the perception aspect. Here we focus on evaluating generated outputs and on the interpretability of the latent space; we leave the question of how the Q-network performs on lexical item identification/discrimination for future work. This brings some limitations in terms of model comparison  --- we lack information on how well a GAN-based unsupervised lexical learner performs on word discrimination tasks compared to, for example, the autoencoder architecture. It is reasonable to assume that GAN-based models would perform worse on word discrimination tasks compared to autoencoders in which both the encoder and decoder have direct access to the training data. The Generator and the Q-network in the proposed architecture do not have a direct access to the training data --- the Generator has only a very indirect access to the data by being trained on maximizing the error of the Discriminator that aims to estimate Wasserstein distance between generated and real data.  Additionally, the fiwGAN and ciwGAN models proposed here have substantially more reduced latent representations (from 5 to 13 variables total).  This likely negatively affects the word discrimination, and as such, the evaluation of discrimination is left for future work.

Another advantage of the GAN architecture is that the networks generate innovative data rather than replicates of data. We can thus probe learning by analyzing how GANs innovate, how they violate data distributions, and what can these innovative outputs tell us about their learning. Additionally,  we focus on exploring how manipulating the latent space (as proposed in \citealt{begus19}) can elicit generation of unique lexical items at categorical levels, what effects individual latent variables have on outputs, and  what this manipulation can tell us about lexical learning in deep convolutional networks.

\subsection{GANs and language acquisition}

The main characteristics of the  GAN architecture \citep{goodfellow14}  are two networks: the Generator and the Discriminator. The Generator generates data from latent space that is reduced in dimensionality (e.g.~from a set of uniformly distributed variables $z$). The Discriminator network learns to distinguish between ``real'' training data and generated outputs from the Generator network. The Generator is trained to maximize the Discriminator's error rate; the Discriminator is trained to minimize its own error rate. In the DCGAN proposal \citep{radford15}, the two networks are deep convolutional networks. Recently, the DCGAN proposal was transformed to model audio data in WaveGAN \citep{donahue19}. The main architecture of WaveGAN is identical to that of DCGAN \citep{radford15}, with the main difference being that the Generator outputs a one-dimensional vector corresponding to time series data (raw acoustic output) and the Discriminator takes one-dimensional acoustic data as its input (as opposed to two-dimensional visual data in DCGAN). WaveGAN also adopts the Wasserstein GAN proposal for a cost function in GANs that improves training \citep{arjovsky17}. Instead of estimating the probability of whether the output is generated or real, WGAN estimates the Wasserstein distance between generated data and real data.

\citet{begus19} models speech acquisition  as a dependency between latent space and generated outputs in the GAN architecture.  The paper proposes a technique for identifying latent variables that correspond to meaningful phonetic/phonological features in the output. The Generator network learns to encode phonetically and phonologically meaningful representations, such as the presence of a segment  in the output, with a subset of variables, i.e.~with reduced representation. Using the technique proposed in \cite{begus19}, we can identify individual variables that correspond to, for example, a sound [s] in the output. By manipulating these identified variables to values that are beyond the training range, we can force [s] in the output. Interpolating the values has an almost linear effect on the amplitude of frication noise of [s] in the output.

One of the advantages of the proposal in \cite{begus19} is that the model  learns phonological alternations, i.e.~context-dependent changes in the realization of speech sounds, simultaneously with learning acoustic properties of human speech. The WaveGAN model \citep{donahue19} is trained on a simple phonological process: aspiration of stops /p, t, k/ conditioned on the presence of [s] in the input. English voiceless stops /p, t, k/ are aspirated (produced with a puff of air [\textipa{p\super h}, \textipa{t\super h}, \textipa{k\super h}]) word-initially before a stressed vowel (e.g.~in \emph{pit} [\textipa{"p\super hIt}]) except if an [s] precedes the stop (e.g.~\emph{spit} [\textipa{"spIt}]) A computational experiment suggests that the network learns this distribution, but imperfectly so.   The network learns to output shorter aspiration duration when [s] is present, in line with distributions in the training data. Outputs, however, also violate data in a manner that can be directly paralleled to language acquisition. Occasionally, the Generator network outputs aspiration durations that are longer in the [s] condition than in any example in the training data: the generator outputs [\textipa{sp\super hIt}], which violates the phonological rule in English. In other words, the network violates the distributions in the training data, and these violations correspond directly to phonological acquisition stages: children acquiring English start with significantly longer aspiration durations in the [s]-condition, e.g.  [\textipa{sp\super hIt}] \citep{bond80}.

In sum, GANs have been shown to represent phonetically or phonologically meaningful information in the latent space that has approximate equivalent in phonetic/phonological representations and language acquisition \citep{begus19}. The latent variables can be actively manipulated to generate data with or without some phonetic/phonological property. These representations, however, are exclusively limited to the phonetic/phonological level in \cite{begus19} and contain no lexical information.

\subsection{Goals}

Despite several advantages, to our knowledge, lexical learning has not yet been modeled with  Generative Adversarial Neural network models and perhaps even more generally, with unsupervised generative deep \emph{convolutional} networks. \cite{donahue19} train the WaveGAN architecture on Speech Commands Zero Through Nine (SC09) dataset and argues that the network learns to generate speech-like outputs with high naturalness and inception scores. \cite{donahue19}, however,  do not explore internal representations and their architecture does not include the Q-network (InfoGAN; \citealt{chen16}), which means the proposal does not model lexical learning or how the network encodes linguistically meaningful representations.  As discussed in Section \ref{prior}, most other models operate with recurrent neural networks rather than with convolutional networks and use the autoencoder architecture.\footnote{A deep convolutional autoencoder model architecture based on WaveNet proposed in \cite{chen20} was released after submission of this paper.}

In this paper, we follow the proposal in \cite{begus19} that phonetic and phonological acquisition can be modeled as a dependency between latent space and generated data in the GAN architecture and add a lexical learning component to the model. We  modify the WaveGAN architecture and add the InfoGAN's Q-network (based partially on  implementation in \citealt{signnet}) to computationally simulate lexical learning from raw acoustic data. In other words, we introduce a deep convolutional network that learns to retrieve the Generator's latent code and propose a new latent space structure that can model featural learning (fiwGAN). The fiwGAN architecture additionally allows a very low dimension categorical vector representation of lexical items (e.g.~$n$ number of features allows $2^n$ number of unique classes). We train the networks on highly variable training data: manually sliced lexical items from the TIMIT database \citep{timit} that includes over 600 speakers from different dialectal backgrounds in American English. We present four computational experiments: on five lexical items in the ciwGAN architecture (Section \ref{ciw5}), on ten lexical items in  the ciwGAN architecture (Section \ref{ciw10}), on eight lexical items in the fiwGAN architecture (Section \ref{fiwGANon8}), and on the entire TIMIT database (6,229 lexical items) in the fiwGAN architecture (Section \ref{fiwGANonTimit}). Evidence for lexical learning emerges in all four experiments. The paper also features a section describing how to directly follow learning strategies of the Generator network (Section \ref{19244}), a section on featural learning that discusses innovative outputs and productivity of the model (Section \ref{fl}), and a section that proposes a technique for retrieving near categorical underlying representation of the latent variables in GANs (Section \ref{up}). We argue that exploration of innovative outputs and the latent space of deep neural networks trained on dependencies on speech data that are well understood (due to extensive study of human phonetics and phonology in the past decades) provides unique insights both for cognitive modeling and for neural network interpretability.

 Lexical learning is modeled in the following way: a deep convolutional network learns to retrieve information from innovative outputs generated  by a separate Generator network. The Generator network thus learns to generate data such that unique lexical information is retrievable from its acoustic outputs. Lexical learning is not \emph{per se} incorporated in the model: instead, lexical learning emerges because the most informative way to generate outputs given speech data as input is to encode unique information into lexical items. The end result of the model is a Generator network that generates innovative data --- raw acoustic outputs --- such that each lexical item is represented by a unique code. Because the model diverges substantially from existing proposals of lexical learning, we leave direct comparison of its performance (such as on ABX test) for future work. Instead, we propose to evaluate success in the model's performance in lexical learning with an inferential statistical technique --- multinomial logistic regression (Section \ref{ciw5}). 
 
 Representing semantic information can take many forms in computational models. In the current proposal, unique lexical items are represented with either a one-hot vector in the ciwGAN architecture or a binary code in the fiwGAN architecture. In other words, the objective of the model is to associate each unique lexical item in the training data with a unique representation. For example, in a corpus with four words, \emph{word1} can be associated with a representation [1, 0, 0, 0], \emph{word2} with  [0, 1, 0, 0], \emph{word3} with [0, 0, 1, 0] in the ciwGAN architecture. In the fiwGAN architecture, \emph{word1} can be associated with [0, 0], \emph{word2} with [0, 1], \emph{word3} with [1, 0], and \emph{word4} with [1, 1].

The model of lexical learning proposed here features some desirable properties. First, the network is trained exclusively on raw unannotated acoustic data. Second, lexical learning emerges from the requirement on a deep convolutional network to output informative data. Only because associating a unique code in the latent space with lexical items is the optimal way to encode information such that another network will be able to retrieve it does the lexical learning emerge. 
Third, the model  is fully generative: a deep convolutional network (the Generator) generates raw acoustic outputs that correspond to lexical items in the training data.  Crucially, the Generator network in the model does not simply replicate training data, but generates innovative outputs, because its main task is to increase the error rate of the network that distinguishes real from generated data (the Discriminator) and its outputs are not directly connected to the training data.  Occasionally, the Generator outputs innovative data that violate distributions of the training data, but are linguistically interpretable and highly informative. The model thus features one of the basic properties of language: productivity. This allows us to compare lexical and phonological acquisition in language-acquiring children to the innovative generated data in the proposed computational model. The fiwGAN architecture has an additional advantage: it can model featural learning in addition to a higher level classification. This means that featural representations in phonology and phonetics can be modeled simultaneously with lexical learning.

To be sure, there are also undesirable aspects of the model. In particular, the model's performance is optimal when the number of lexical classes that the network is predetermined. However, as the experiment in Section \ref{fiwGANonTimit} suggests, even with the mismatch between the number of classes and the number of unique items, the networks show evidence for lexical learning. Also, while the model learns from raw acoustic inputs, the individual lexical items in training data are sliced from the corpus (sliced at the lexical level rather on the phone level) instead of inferred by the model. These disadvantages are not insurmountable, but are left to be addressed in future work.

The proposed architectures and results of the computational experiments have implications for deep neural network interpretability as well as some basic implications for NLP applications.   Beside modeling lexical learning, the novel latent space structure in the fiwGAN architecture can be employed as a general purpose unsupervised simultaneous feature extractor and classifier for audio data. We also propose a technique for exploring latent space representations: we argue that manipulating latent codes to marginal values that substantially excess the training range reveals underlying values for each latent code. Outputs generated with the proposed technique feature little variability and have the potential to reveal learning representations of the Generator network. The proposed model also allows a first step towards unsupervised text-to-speech synthesis at the lexical level using GANs: the Generator outputs specific lexical items when latent codes are set to different values.

\section{Model}

We propose a GAN architecture that combines WaveGAN with the InfoGAN proposal \citep{chen16}.  
The objective of Generative Adversarial Networks is a function that maps from randomly-distributed latent space to outputs that resemble training data (the Generator network). To find such a function, the Generator and the Discriminator networks are trained in a minimax game in which the Discriminator's (D) loss is maximized and the Generator's (G) loss is minimized \citep{goodfellow14}. In the original GAN proposal \citep{goodfellow14}, the Discriminator is trained on classifying real and fake data. In the Wasserstein GAN proposal that is adopted in this paper (as well as in \citealt{donahue19}) and that substantially improves training, the Discriminator is trained on minimizing the Wasserstein distance between the generated and real data distributions \citep{arjovsky17}. The value function can be formalized as \citep{arjovsky17,donahue19}:

\begin{equation}\label{gans0}
V_{WGAN}(D, G) =  \mathbb{E}_{x\sim P_{X}}[D_w(x)] - \mathbb{E}_{z\sim P_Z}[D_w(G(z))],
\end{equation}

where $x$ is real data from some data distribution ($P_{X}$) and $z$ is latent space from a random distribution ($P_Z$; in our case the uniform distribution). To improve performance, the models are additionally trained with a gradient penalty term $\lambda\mathbb{E}_{\hat{x}\sim P_{\hat{x}}}[(||\triangledown_{\hat{x}} D_w(\hat{x})||_2-1)^2]$, where  $P_{\hat{x}}$ is a uniform probability distribution in the interval [0, 1] which is used to sample from the difference between the real and generated data distributions ($\hat{x}$) to get the gradient penalty and  $\lambda$ is a constant set at 10 (for advantages of such a gradient penalty term over weight clipping, see the WGAN-GP proposal in \citealt{gulrajani17}).

InfoGAN \citep{chen16} is a proposal within the GAN framework that aims to increase mutual information between a subset of latent space --- the code variables ($c$ or $\phi$) --- and generated outputs ($G(z, c)$) \citep{chen16}.  In this paper,  we adopt the main objectives from the Wasserstein proposal (\ref{gans0}) and add to the model maximization of mutual information between the code variables in the latent space and the generated outputs $I(c; G(z,c))$. Because $I(c; G(z,c))$ is difficult to estimate, \cite{chen16} instead propose to approximate its variational lower bound  $\lambda L_I (G, Q)$ (with a hyperparameter $\lambda$; for details, see \citealt{chen16}).  Our model can thus be formalized as (based on \citealt{chen16} and \citealt{gulrajani17};  see also \ref{gans0}):

\begin{equation}\label{gans1}
\min_{G,Q}\max_{D} V_{IWGAN}(D, G, Q) = V_{WGAN}(D, G) -\lambda L_I (G, Q).
\end{equation}

To implement this model, the proposed ciwGAN and fiwGAN architectures involve three deep convolutional networks: the Generator, the Discriminator, and the Q-network (or the lexical learner). The models are based on WaveGAN \citep{donahue19}, an implementation of the DCGAN architecture \citep{radford15} for audio data and the InfoGAN proposal \citep{chen16}.\footnote{\cite{barry19} in a recent presentation model piano playing with InfoWaveGAN. Their proposal, however, focuses on continuous variables and feature only one categorical latent variable with no apparent function. It is unclear from the poster what the architecture of their proposal is. }  Unlike in most InfoGAN implementations, the Q-network is a separate deep convolutional network.\footnote{The InfoGAN model based on DCGAN in \cite{signnet} also proposes the Q-network to be a separate network.} 

\begin{figure}
\centering
\includegraphics[width=0.8\textwidth]{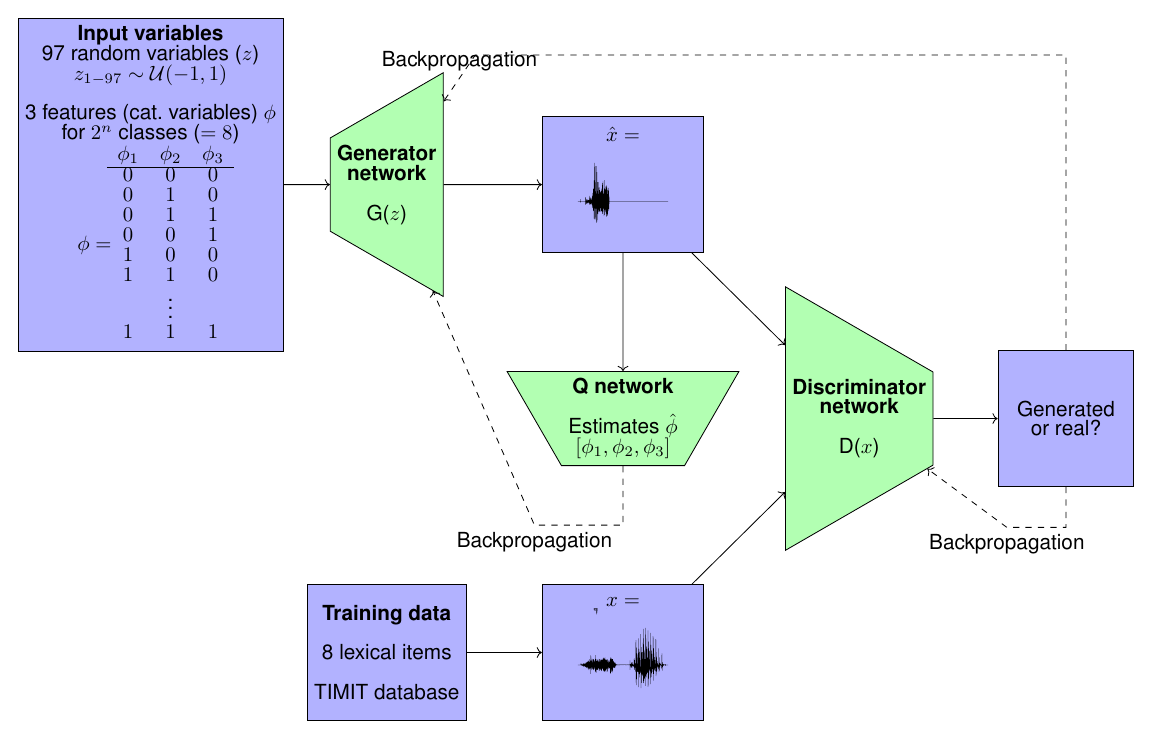}

\caption{\label{fiwgantikz}Architecture of fiwGAN: green trapezoids represent deep convolutional neural networks; purple squares illustrate inputs to each of the three networks. The Generator network takes 3 latent features $\phi$ (constituting binary code) and 97 latent variables $z$ uniformly distributed ($z\sim\mathcal{U}(-1,1)$) as its input. The Generator outputs a vector of 16384 values ($\hat{x}$) that constitute approximately 1 s of audio file (sampled at 16000 Hz). The Discriminator takes generated data ($\hat{x}$) and real data and estimates the Wasserstein distance between them. The Q-network (lexical learner) takes generated data as its input and outputs estimates of the unique feature values that the Generator uses for generation of each data point. }
\end{figure}

In the GAN architecture, the Generator network usually takes as its input a number of uniformly distributed latent variables ($z\sim\mathcal{U}(-1,1)$). In the InfoGAN proposal \citep{chen16}, the Generator's input additionally includes a latent code: a set of binary variables that constitutes a one-hot vector as well as uniformly distributed code variables. Because we model lexical learning, we exclude uniformly distributed code variables. While the binary variables in InfoGAN implementations usually constitute a one-hot vector, we propose two different architectures. The ciwGAN architecture includes a one-hot vector as its latent code ($c$); but in the fiwGAN architecture, we introduce binary code as the categorical input (labeled as $\phi$).\footnote{For a different kind of binarization that applies to the entire latent space in the variational autoencoder architecture, see \cite{eloff19}.} This new structure in the fiwGAN latent space allows the network to treat the binary variables as features, where each variable corresponds to one feature ($\phi_n$). As a consequence, the two networks differ in how the Q-network is trained. In ciwGAN, the Q-network is trained on retrieving information from the Generator's output with a softmax function in its final layer. In fiwGAN, the categorical variables or features are binomially distributed and the Q-network is trained to retrieve information with a sigmoid function in the final layer accordingly.  In sum, the Generator in our proposal takes two sets of variables as its input (latent space): (i) categorical variables ($c$ or $\phi$) which constitute a one-hot vector (ciwGAN) or a binary code (fiwGAN) and (ii) random variables $z$ that are uniformly distributed ($z\sim\mathcal{U}(-1,1)$). Figure \ref{fiwgantikz} illustrates the fiwGAN architecture.  The code is available at \url{https://github.com/gbegus/fiwgan-ciwgan}.

The  Generator network is a five-layer deep convolutional network (from  WaveGAN; \citealt{donahue19}) that takes the input variables (referred to as the latent variables or the latent space) and outputs a 1D vector of 16,384 data points that constitute just over 1 second of acoustic audio output with 16 kHz sampling rate. These generated outputs are fed to the Discriminator network and the Q-network. The Discriminator network takes raw audio as its input: both generated data and real data sliced at the lexical level from the TIMIT database \citep{timit}. It is trained on estimating the  Wasserstein distance between generated and real data distributions, according to \cite{arjovsky17}. It outputs ``realness'' scores which estimate how far from the real data distribution an input is \citep{brownlee19}. The Generator's objective is to increase the error rate of the Discriminator such that the Discriminator assigns a high ``realness'' score to its generated outputs.

To model lexical learning, we add the Q-network  to the architecture (InfoGAN; \citealt{chen16}). As already mentioned, the Q-network is independent of the Discriminator network in the proposed architecture.  A separate Q-network is in fact desirable as it enables exploration and probing of internal representations  that are limited to lexical learning and are dissociated from the function of the Discriminator. In future work, we can thus test the Q-network on discriminative tasks (such as ABX) and probe its representations that are limited to lexical learning and are not influenced by the Discriminator's function (of estimating realness scores). The Q-network is in its architecture identical to the Discriminator. It takes only generated outputs ($G(z)$) as its input in the form of 16384 data points (approximately 1 s of audio data sampled at 16 kHz). The Q-network has 5 convolutional layers. The only difference between the Discriminator and the Q-network is that the final layer in the Q-network includes $n$ number of nodes, where $n$ corresponds to the number of categorical variables ($c$ in ciwGAN) or features ($\phi$ in fiwGAN) in the latent space.

The Q-network is trained on estimating the categorical part of the latent space ($c$- or $\phi$-values). Its output is thus a unique code that approximates the latent code in the Generator's latent space --- either a one-hot vector or a binary code. The training objective of the Q-network is to approximate the unique latent code in the Generator's hidden input. The loss function of the Q-network is to minimize the difference between the estimated $c$- or $\phi$-values that correspond to the nodes in the last fully connected layer and the actual $c$- or $\phi$-values in the latent space of the Generator. At each evaluation, weights of the Q-network as well as the Generator network are updated with cross-entropy according to the loss function of the Q-network. This forces the Generator to generate data, such that the latent code or latent features ($c$ or $\phi$) will be retrievable: the Generator's objective is to maximize the success rate of the Q-network. The Generator is additionally trained on maximizing the error rate of the Discriminator. The training proceeds as follows: the Discriminator network is updated five times, followed by an update of the Generator based on the Discriminator's loss and an update of the Generator together with the Q-network to minimize the Q-network's loss.  The Generator and the Discriminator networks are trained with the Adam optimizer, whereas the Q-network is trained with the RMSProp algorithm (with the learning rate set at .0001 for all optimizers).  The minibatch size is 64.

To summarize the architecture, the Discriminator network learns to distinguish ``realness'' of generated speech samples. The Generator is trained to maximize the loss function of the Discriminator. The Q-network (or the lexical learner network) is trained on retrieving the categorical part of the latent code in the Generator's output based on only the Generator's acoustic outputs. Because the weights of the Q-network as well as the Generator are updated based on the Q-network's loss function, the Generator learns to associate lexical items with a unique latent code (one-hot vector or binary code), so that the Q-network can retrieve the code from the acoustic signal only. This learning that resembles lexical learning is  unsupervised: the association between the code in the latent space and individual lexical items arises from training and is not pre-determined. In principle, the Generator could associate any acoustic property with the latent code, but it would make it harder for the Q-network to retrieve the information if the Generator encoded some other distribution with its latent code. The association between a unique code value and individual lexical item that the Generator outputs thus emerges from the training.

The result of the training in the architecture outlined in Figure \ref{fiwgantikz} is a Generator network that outputs raw acoustic data that resemble real data from the TIMIT database, such that the Discriminator becomes unsuccessful in assigning ``realness'' scores \citep{brownlee19}. Crucially, the Generator's outputs are never a full replication of the input: the Generator outputs innovative data that
 resemble input data, but also violate many of the distributions in a linguistically interpretable manner \citep{begus19}. In addition to outputting innovative data that resemble speech in the input, the Generator also learns to associate each lexical item with a unique code in its latent space. This means that by setting the code to a certain value, the network should output a particular lexical item to the exclusion of other lexical items.

There are two  supervised aspects of the model. First, the network is trained on manually sliced lexical items and does not perform slicing from a continuous speech stream in an unsupervised manner. Addressing this disadvantage is left for future work (see the work on this topic in \citealt{lee15,rasanen15,rasanen20}). Second, the model performs best when the number of lexical items in the training data matches the number of classes predetermined in the model.  For example, a one-hot vector in the ciwGAN architecture with 5 variables is used to categorize 5 lexical items. We feed the network with 5 different lexical items from the TIMIT database. In the fiwGAN architecture, $n$ features ($\phi$) are used to categorize $2^n$ classes. For example, 3 features $\phi$ allow $2^3 = 8$ classes and we feed the network 8 different lexical items. However, as is suggested by  the experiment in Section \ref{fiwGANonTimit}, the Generator learns to associate single lexical items with a given latent code even if there is a high mismatch between the number of classes and the number of lexical items.  In other words, that the number of classes and actual lexical items match is not a hard requirement and evidence for lexical learning emerges even if the number of possible classes is substantially higher than the number of actual items. One disadvantage in such a case is that occasionally high frequency words can be associated with multiple codes (see discussion in Section \ref{fiwGANonTimit}).

\section{Experiments}

\subsection{ciwGAN on 5 lexical items}
\label{ciw5}

\subsubsection{8011 steps}
\label{8011}

The first model is trained on the ciwGAN architecture with 5 lexical items from TIMIT: \textit{oily}, \textit{rag}, \textit{suit}, \textit{water}, and \textit{year}. The latent space of this network includes 5 categorical variables ($c$) constituting a five-level one-hot vector and 95 random variables $z$. The five lexical items were chosen based on frequency: they are chosen from the most frequent content words  with at least 600 data points in TIMIT. A total of 3205 data points were used in training and each of the five items has $>600$ data points in the training data.  The input data are 16-bit .wav slices of lexical items (as annotated in TIMIT) sampled with 16 kHz rate. Input lexical items with counts are given in Table \ref{5ciwWords}.

\begin{table}\centering
\begin{tabular}{llc}\hline\hline
word&IPA&data points\\
\hline
oily&\textipa{["OIli]}& 638\\
rag &\textipa{["\*r\ae g]}&638\\
suit &\textipa{["sut]}&630\\
water &\textipa{["wOR\textrhookschwa]}&649\\
year &\textipa{["jI\*r]}&650\\\hline
\textbf{Total}&&3205\\
\hline\hline
\end{tabular}
\caption{\label{5ciwWords}Five lexical items used for training in the five-word ciwGAN model with their corresponding IPA transcription (based on general American English) and counts of data points for each item.}

\end{table}

To test whether the Generator network learns to associate each lexical item with a unique code, the ciwGAN architecture is trained  after 8011 ($\sim$ 800 epochs) and  19244 steps ($\sim$ 1921 epochs) and 100 outputs are generated for each one-hot vector. \cite{begus19} shows that manipulating the latent space of the Generator network to values outside of the training interval can reveal the underlying feature encoded with each variable. Additionally,  \cite{begus19} argues that the relationship between the latent variables and meaningful phonetic properties can be almost linear.  Based on these findings, the code variables ($c$) in the generated samples are manipulated not to 1 (as in the training stage), but rather to 2 when generating outputs. The rest of the latent space (all $z$-variables) are sampled randomly, but kept constant across the five categorical variables.

One hundred outputs are thus generated for each unique code (e.g.~[2, 0, 0, 0, 0], [0, 2, 0, 0, 0] ...).\footnote{All acoustic analyses are performed in Praat \citep{boersma15}.} We analyze outcomes at two points during the training: after 8011 steps ($\sim$ 800 epochs) and after 19244 steps ($\sim$ 1921 epochs). Since we are modeling language acquisition, we are not interested in full convergence of the model: it is more informative to probe the network as it is being trained. The number of steps at which we probe the networks is somewhat arbitrary, but the main consideration in choosing the number of steps is  a balance between interpretability of outputs and minimizing the number of epochs (for a more detailed discussion, see \citealt{begus19}). The outputs were analyzed by a phonetically trained female speaker of American English who is not a co-author in this research and was not aware of the exact details of the experiment. The results below are reported based on the transcriber's analysis as well as based on an acoustic analysis by the author.  Altogether, 1000 outputs are thus analyzed and transcribed. 

Results of the analysis suggest that the network associates each unique code with a different lexical item. The success rate, however, differs across  lexical items.  For example, when the latent code is set at [0, 0, 0, 0, 2] the Generator trained after 8011 steps outputs samples that are transcribed as \emph{rag} in 98/100 cases. In other words, the Generator learns to associate [0, 0, 0, 0, 2] with \emph{rag}.\footnote{Occasionally, a short vocalic element precedes the \textipa{["\*r\ae g]}.}  The Generator thus not only learns to generate speech-like outputs, it also represents distinct lexical items with a unique representation: information that can be retrieved from its outputs by the lexical learning network. We can argue that [0, 0, 0, 0, 2] is the underlying representation of \emph{rag}.\footnote{In the remaining two cases, the outputs include [\textipa{\*r}] in the initial position, which is followed by a diphthong [\textipa{aI}] and a period of a consonantal closure. One output was transcribed as \emph{right}.}

\begin{figure}
\centering
\includegraphics[width=.5\textwidth]{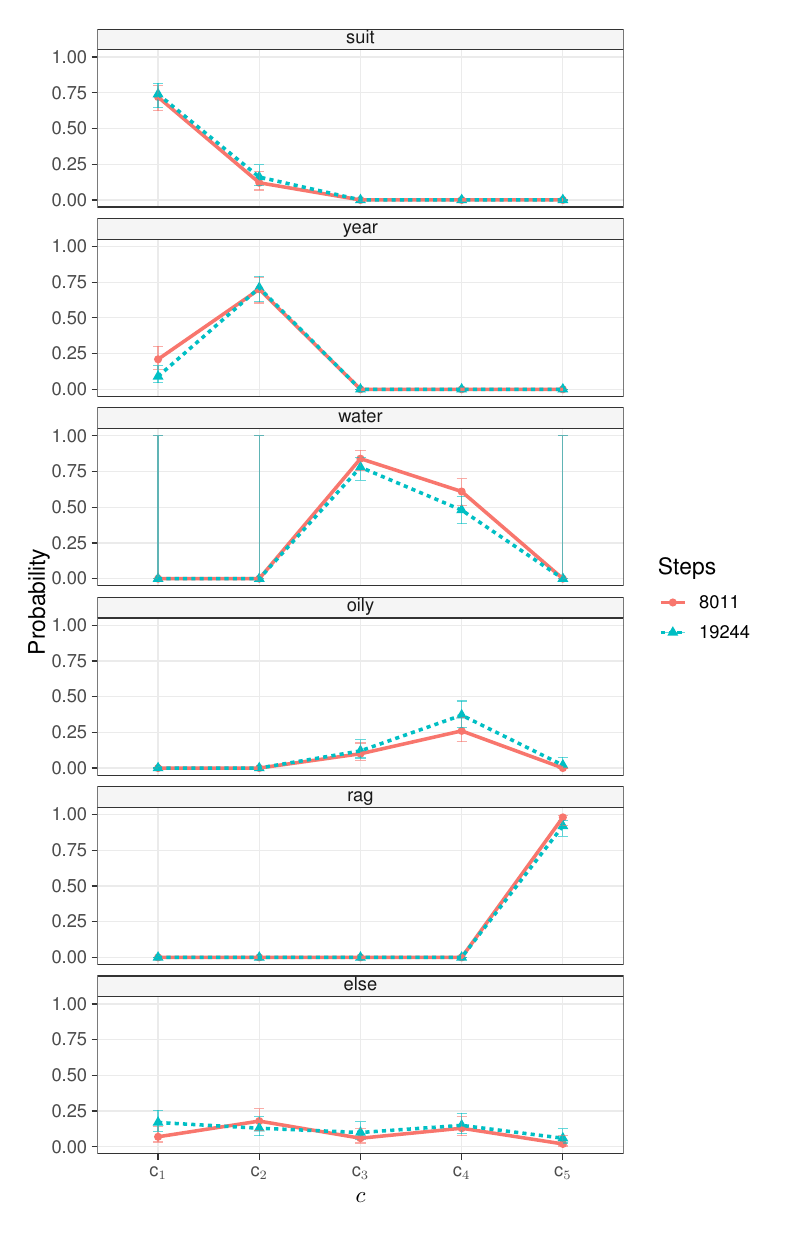}

\caption{\label{ceffect} Estimates of two multinomial logistic regression models (for 5-word models trained after 8011 and 19244 steps) with coded transcribed outputs as the dependent variable and the latent code with five levels that correspond to the five unique one-hot vectors in the model. }
\end{figure}

 To determine the underlying code for each lexical item, we use success rates (or estimates from the multinomial logistic regression model in Table \ref{5counts} and Figure \ref{ceffect}): the lexical item that is the most frequent output for a given latent code is assumed to be associated with that latent code (e.g.~\emph{rag} with [0, 0, 0, 0, 2]). Occasionally, a single lexical item is the most  frequent output for two latent codes. As will be shown below, it is likely the case that this reflects imperfect learning where the underlying lexical item for a latent code is obscured by a more frequent output (perhaps the one that is easier to distinguish from the data).  In this case, we associate such codes to the lexical item for which the given code outputs the highest proportion of that lexical item with respect to other latent codes. For example, [0, 0, 0, 2, 0] outputs \emph{water} most frequently with \emph{oily} accounting for approximately a quarter of outputs. The assumed lexical item for [0, 0, 0, 2, 0] is \emph{oily}, because the code that outputs \emph{water} most frequently is [0, 0, 2, 0, 0], while highest proportion of \emph{oily}  relative to other latent codes is for [0, 0, 0, 2, 0]. Observing the progress of lexical learning provides additional evidence that \emph{oily} is the underlying representation of [0, 0, 0, 2, 0]: as the training progresses the network increases accuracy  (see Section \ref{19244}).

The success rate for the other four lexical items is lower than for \emph{rag}, but the outputs that deviate from the expected values are highly informative. For $c = [2, 0, 0, 0, 0]$, the Generator (8011 steps) outputs 72/100 data points that can be reliably transcribed as \emph{suit}. In seven additional cases, the network outputs data points that can be transcribed with a sibilant [s] (79 total). In these outputs, [s] is followed by a sequence that either cannot reliably be transcribed as \emph{suit} or does not correspond to \emph{suit}, but rather to \emph{year} (transcribed as \emph{sear} [\textipa{sI\*r}]). The remaining 21 outputs do not include the word for \emph{suit} or a sibilant [s]. However, they are not randomly distributed across other four lexical items either --- they  include lexical item \emph{year} or its close approximation.\footnote{For raw counts in this and other models, see Tables \ref{app1},  \ref{app2},  \ref{app3}, and \ref{app4}.}

\begin{figure}
\centering
\includegraphics[width=1\textwidth]{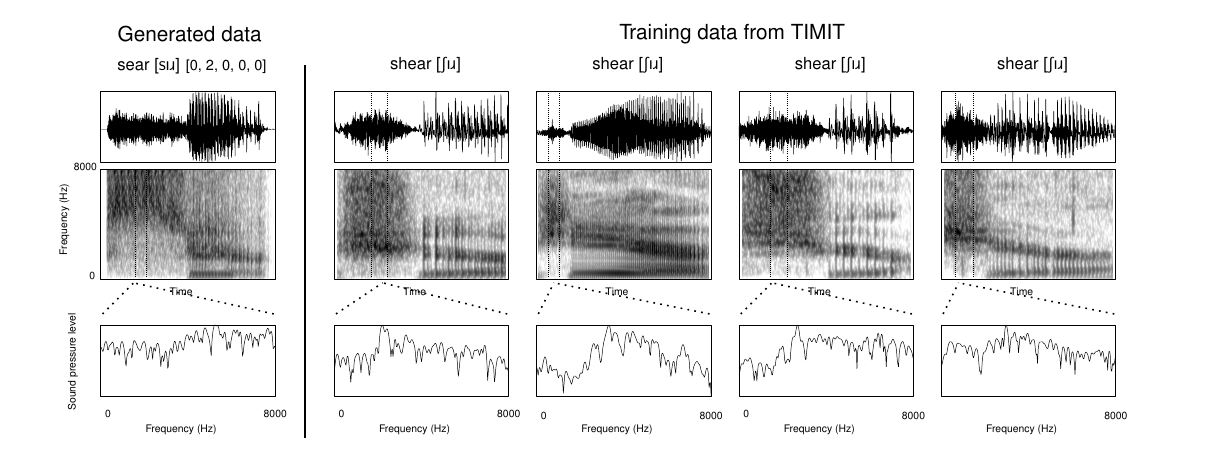}

\caption{\label{searPrava}Waveforms (top), spectrograms (mid, from 0--8000 Hz), and 25 ms spectra (slices indicated in the spectrograms) (bottom) of four data points of the lexical item \emph{year} with clear frication noise in the training data (from TIMIT) and the generated innovative output \emph{sear}. }
\end{figure}

An acoustic analysis of the training data reveals motivations for the innovative deviating outputs. 
 As already mentioned, the network occasionally generates an innovative output, \emph{sear}. The sources of this innovation are likely four cases in the training data in which [j] in \emph{year} ([\textipa{jI\*r}]) is realized as a post-alveolar fricative [\textipa{S}], probably due to contextual influence (something that could be transcribed as \emph{shear} [\textipa{SI\*r}]). Figure \ref{searPrava} illustrates all four examples. The innovative generated output \emph{sear} differs from the four examples in the training data in one crucial aspect: the frication noise in the generated output is that of a post-alveolar [s] rather than that of a palato-alveolar [\textipa{S}]. Spectral analysis in Figure \ref{searPrava} clearly shows that the center of gravity in the generated output is substantially higher than in the training data (which is characteristic of the alveolar fricative [s]).

The innovative \emph{sear} output likely results from the fact that the training data contains four data points that pose a learning problem: \emph{shear} that features elements of \emph{suit} and \emph{year}. The innovative generated  \emph{sear} [\textipa{sI\*r}] consequently features (i) frication noise that is  approximately consistent with \emph{suit} [\textipa{sut}]  and (ii) formant structure consistent with  \emph{year} [\textipa{jI\*r}]. It appears that the network treats  \emph{sear} as a combination of the two lexical items. The network generates innovative outputs that combines the two elements (\emph{sear} [\textipa{sI\*r}]). Additionally, the \emph{sear} output seems to be equally distributed among the two latent codes, [2, 0, 0, 0, 0] representing \emph{suit} and [0, 2, 0, 0, 0] representing \emph{year}. In other words, the error rate distribution of the two latent codes suggests that the network classifies the output \emph{sear} as the combination of elements consistent with [2, 0, 0, 0, 0]  and [0, 2, 0, 0, 0].

For $c = [0, 2, 0, 0, 0]$, the Generator   outputs 68 data points that can be reliably transcribed as \emph{year} or at least have a clear [\textipa{I\*r}] sequence  (without an [s]).\footnote{The initial consonant is sometimes absent from transcriptions, but this is primarily because the glide interval is acoustically not prominent, especially before [\textipa{I}].} 22 outputs feature a sibilant [s]. In these 22 cases, 16 can reliably be transcribed as \emph{suit}, while the others are mostly variants of the innovative \emph{sear}. The remaining cases (approximately 10) are difficult to categorize based on acoustic analysis.

\begin{figure}
\centering
\includegraphics[width=0.8\textwidth]{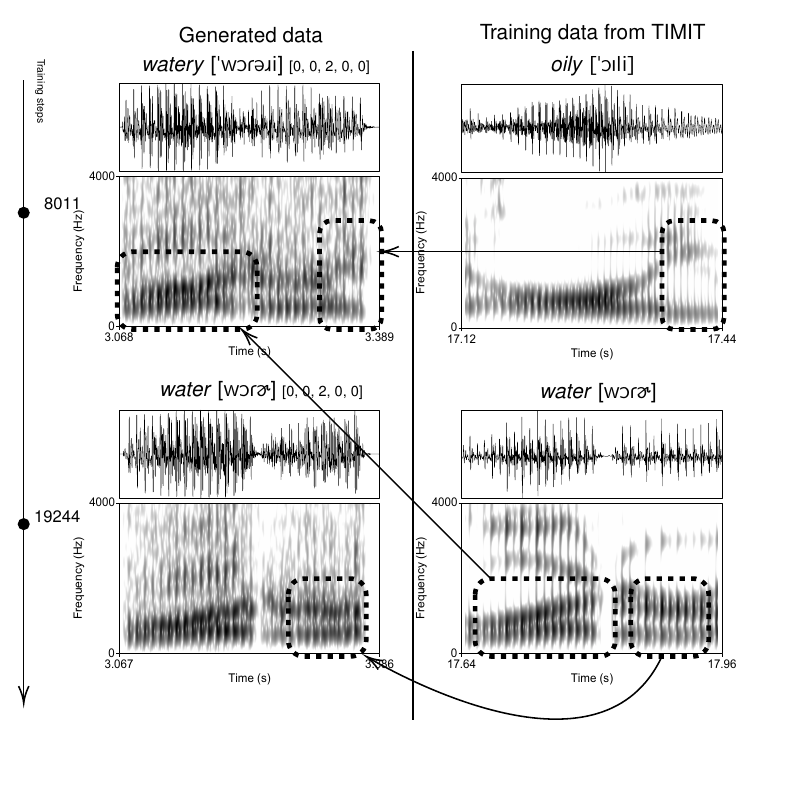}

\caption{\label{watery}A waveform and spectrogram (0--4000 Hz) of an innovative output \emph{watery} [\textipa{"wOR\textschwa\*ri}] (top right). That innovative \emph{watery} is a combination of \emph{water} \textipa{["wOR\textrhookschwa]} and \emph{oily}  \textipa{["OIli]} is illustrated by two examples from the training data (top and bottom right). The innovative output \emph{watery} features a clear formant structure of \emph{water} with a high front vowel [i], characteristic of the lexical item \emph{oily} (see marked areas of the spectrograms). At 19244 steps, the vocalic structure of [i] is not present in the output, given the exact same latent code and random latent space. The network thus corrects the formant structure from an innovative \emph{watery} into \emph{water} [\textipa{"wOR\textschwa\*r(\textschwa)}] as the training progresses. In some other cases, the network at 19244 steps outputs \emph{oily} for what was \emph{watery} at 8011 steps.}
\end{figure}

For [0, 0, 2, 0, 0], the Generator outputs 84 data points that are transcribed as containing \emph{water} [\textipa{wOR\textrhookschwa}]. In approximately 15 of the 84 cases, the output involves an innovative combination transcribed as \emph{watery} [\textipa{"wOR\textschwa\*ri}]. Figure \ref{watery} illustrates one such case.   \emph{Watery} is an innovative  output that combines segment [i] from \emph{oily} (\textipa{["OIli]}) with [\textipa{"wOR\textschwa\*r}] from \emph{water} into a linguistically interpretable innovation. This suggests that the Generator outputs a novel combination of segments, based on analogy to \emph{oily}. Unlike for \emph{sear}, the training data contained no direct motivations based on which \emph{watery} could be formed.\footnote{In 10 further cases of [0, 0, 2, 0, 0], the Generator outputs data points that contain a sequence \emph{oil} \textipa{["OIl]}.  Transcription of the remaining 6 outputs is uncertain.}

Finally, for [0, 0, 0, 2, 0], the Generator outputs only 26 outputs that can reliably be transcribed as \emph{oily} \textipa{["OIli]}. On the other hand, 61 outputs contain \emph{water}.  \emph{Oily} is the less frequent output for [0, 0, 0, 2, 0] compared to \emph{water}, but \emph{water} is assigned to [0, 0, 2, 0, 0]  because it is its most frequent output, while [0, 0, 0, 2, 0] is the code that outputs the highest proportion of \emph{oily}. This is why we analyze \emph{oily} as the underlying lexical item for the [0, 0, 0, 2, 0] code. Another  evidence that \emph{oily} might underly the [0, 0, 0, 2, 0] code is that as the training progresses, the Generator increases the number of outputs transcribed with \emph{oily} for this code and decreases the number of outputs \emph{water} for the same code (see Section \ref{19244} and Figure \ref{ceffect}). For a confirmation that the proposed method for assigning underlying assumed words for a given code based on annotated outputs yields valid results, see Section \ref{up}.

\begin{table}\centering
\begin{tabular}{cl|lr|lr|r}\hline\hline
Assumed word&Latent code $c$&Most frequent&\%&2nd most freq.&\%&Else\\
\hline
suit &[2, 0, 0, 0, 0]&\emph{suit} \textipa{["sut]}&72\%&\emph{year} \textipa{["jI\*r]}&21\%&7\%\\
year&[0, 2, 0, 0, 0]&\emph{year} \textipa{["jI\*r]}&70\%&\emph{suit} \textipa{["sut]}&12\%&18\%\\
water & [0, 0, 2, 0, 0]&\emph{water}\textipa{["wOR\textrhookschwa]}&84\%&\emph{oily} \textipa{["OIli]}&10\%&6\%\\
oily& [0, 0, 0, 2, 0]&\emph{water} \textipa{["wOR\textrhookschwa]}&61\%&\emph{oily} \textipa{["OIli]}&26\%&13\%\\
rag &[0, 0, 0, 0, 2]&\emph{rag} \textipa{["\*r\ae g]}&98\%&|&|&2\%\\

\hline\hline
\end{tabular}
\caption{\label{5counts}Generated outputs and their percentages across the five one-hot vectors in the latent code. Transcriptions of the outputs were coded as detailed in footnote \ref{fntr}. }

\end{table}

To evaluate lexical learning in the ciwGAN model statistically, we analyze the results with a multinomial logistic regression model. To test significance of the latent code as the predictor of the lexical item, annotations of the generated data were coded and fit to a multinomial logistic regression model using the \emph{nnet} package \citep{nnet} in \cite{R}. The dependent variable is the transcriptions of the generated outputs for the five lexical items and the \emph{else} condition.\footnote{\label{fntr}The following conditions were used for coding the transcribed output: if the annotator transcribed an output as containing ``suit'', the coded lexical item was \emph{suit}), if ``ear'' or ``eer'' (and no ``s'' immediately preceding), then year, if involving ``water'', ``oil'', and ``rag'', then \emph{water}, \emph{oily}, \emph{rag}, respectively. In all other cases, the output was coded as \emph{else}.} The independent variable is a single predictor: the latent code with the five levels that correspond to the five unique one-hot values in the latent code. The difference in AIC between the model with the latent code as a predictor ($AIC=674.7$)  vs.~the empty model ($AIC = 1707.1$) suggests that the latent code is indeed a significant predictor of the lexical item in the output. Counts are given in Table \ref{5counts}. Estimates from the multinomial logistic model in Figure \ref{ceffect} clearly show that each lexical item is associated with a unique latent code.

\subsubsection{19244 steps}
\label{19244}

The proposed model of lexical learning allows not only for the ability to test learning of lexical items, but also to probe learning representations as training progresses. We propose that the progress of lexical learning can be directly observed by keeping the random variables constant across training steps. In other words, we train the Generator at various training steps and generate outputs for models trained after different number of steps with the same latent code ($c$) and the same latent variables ($z$). This reveals how encoding of lexical items with unique latent codes changes with training.

To probe lexical learning as training progresses, we train the 5-word model at 8011 steps for an additional 11233 steps (total 19244) and generate outputs. The generation is performed as described in Section \ref{8011}: for each unique latent code (one-hot vector), we generate 100 outputs with latent variables identical to the ones used on the model trained after 8011 steps. The latent code is again manipulated to value 2 (e.g.~[2, 0, 0, 0, 0]) in order to probe the underlying effects of the latent code on generated outputs.

The latent code remains a significant predictor in a multinomial logistic regression model (AIC = 759.9 for a model with the predictor and 1760.2 for an empty model). In fact, success rates remain almost identical across the training steps as is clear from regression estimates in Figure \ref{ceffect} with one notable exception. The most substantial improvement in success rate is observed for \emph{oily}: +11\% in raw counts. The overall success rate is lowest in the 8011-step model precisely for lexical item \emph{oily}. In fact, the success rate for [0, 0, 0, 2, 0] with assumed lexical representation \emph{oily} is only 26\%. At 19244 steps, the success rate (given the exact same latent variables) increases to 37\%.\footnote{The model does not seem to  improve further after 46002 steps.}

\begin{figure}
\centering
\includegraphics[width=0.8\textwidth]{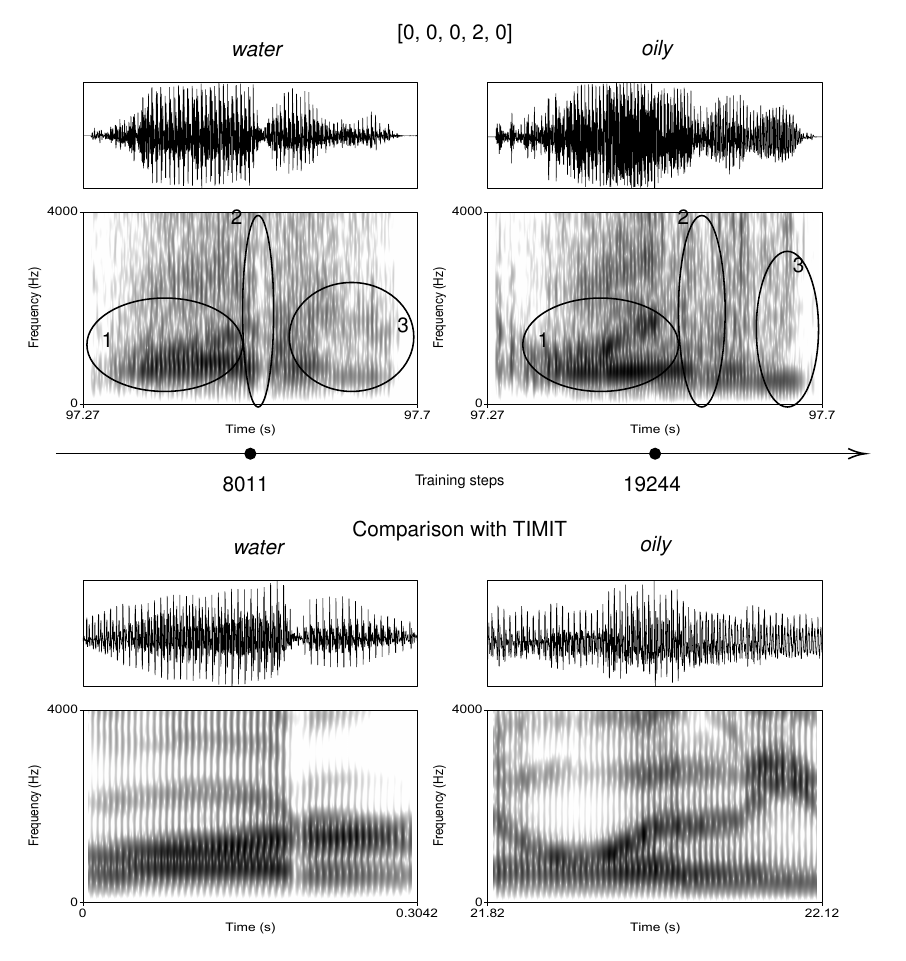}

\caption{\label{wateroilystand} Waveforms and spectrograms (0-4000 Hz) of a generated output at 8011 steps (trained on five lexical items) for latent code [0, 0, 0, 2, 0] that can be transcribed as \emph{water} (top left); and of a generated output for the exact same latent code as well as other 95 latent variables, but generated by a model trained after 19244 steps (top right) transcribed as \emph{oily}. Circled areas point to three major changes on the spectrogram that occur from the output at 8011 steps to the output at 19244 steps: vocalic formants change from [\textipa{wO}] to [\textipa{oI}] (area 1), periods characteristic of a flap [\textipa{R}] change to [l] (area 2) and formant structure for [\textipa{\textrhookschwa}] turns into an [i] (area 3). Examples for \emph{water} and \emph{oily} from the TIMIT database (bottom left and right) illustrate close similarity of the generated outputs to the training data. While the opposite change (from \emph{oily} to \emph{water}) also occurs, it appears less common.}
\end{figure}

Generating data with identical latent variables allows us to observe how the network transforms an output that violates the underlying lexical representation to an output that conforms to it. Figure \ref{wateroilystand} illustrates how an output \emph{water} at 8011 steps for latent code [0, 0, 0, 2, 0] changes to \emph{oily} at 19244 steps.\footnote{Occasionally, a change in the opposite direction is also present.} Both outputs have the same latent code and latent variables ($z$). Spectrograms in Figure \ref{wateroilystand} clearly show how the formant structure of \emph{water} and its characteristic period of reduced amplitude for a flap [\textipa{R}] change to a formant structure characteristic for \emph{oily} with a consonantal period that corresponds to [l]. The figure also features spectrograms of two training data points, \emph{water} and \emph{oily}, which illustrate a degree of acoustic similarity between the two lexical items. Similarly, Figure \ref{watery} illustrates how an  output \emph{watery} that violates the training data in a linguistically interpretable manner at 8011 steps changes to \emph{water} consistent with the training data.

In sum, the results of the first model suggest that the Generator in the ciwGAN architecture trained on 5 lexical items learns to generate innovative data such that each unique latent code corresponds to a lexical item. In other words, the network encodes unique lexical information in its acoustic outputs based solely on its training objective: to generate data such that unique code is retrievable from its outputs.  Figure \ref{ceffect} illustrates that each lexical item is associated with a unique code. Modeling of lexical learning is thus fully generative: when the latent code is manipulated outside of the training range to value 2, the network mostly outputs one lexical item per unique code with success rates  from approximately 98\% to 26\%. The errors are not randomly distributed: the pattern of errors as well as innovative outputs suggest that  (i) \emph{suit} and \emph{year} and (ii) \emph{water} and \emph{oily} are the items that the Generator associates more closely together.  Output errors fall almost exclusively within these groups. The Generator also outputs innovative data that violate training data distributions. Acoustic analysis of the training data reveals motivations for the innovative outputs.  When we follow learning across different training steps, we observe the Generator's repair of innovative outputs or outputs that deviate from the expected values. The highest improvement is observed in the lexical item with overall highest error rate.

\subsection{ciwGAN on 10 lexical items}
\label{ciw10}

To evaluate how the Generator performs on a higher number of lexical classes, another model was trained on 10 content lexical items from the TIMIT database, each of which is attested at least 600 times in the database. All 10 lexical items with exact counts and IPA transcriptions are listed in Table \ref{10items}. 

\begin{table}\centering
\begin{tabular}{llc}\hline\hline
word&IPA&data points\\
\hline
ask&[\textipa{"\ae sk}]&633\\
carry&[\textipa{"k\super h\ae\*ri}]&632\\
dark&[\textipa{"dA\*rk}]&644\\
greasy&[\textipa{"g\*risi}]&630\\
like&[\textipa{"laIk}]&697\\
oily&\textipa{["OIli]}& 638\\
rag &\textipa{["\*r\ae g]}&638\\
suit &\textipa{["sut]}&630\\
water &\textipa{["wOR\textrhookschwa]}&649\\
year &\textipa{["jI\*r]}&650\\\hline
\textbf{Total}&&6441\\
\hline\hline
\end{tabular}
\caption{\label{10items}Ten content lexical items from the TIMIT database used for training in the ciwGAN model (Section \ref{ciw10}) with their corresponding IPA transcription (based on general American English) and counts of data points for each item.}

\end{table}

To evaluate lexical learning in a generative fashion, we use the same technique as on the 5-item Generator in Section \ref{ciw5}. The Generator is trained for 27103 steps ($\sim$ 1346 epochs). The number of epochs is thus approximately at the halfway point between the models in Sections \ref{8011} (8011) and \ref{19244} (19244). We generate 100 outputs for each unique one-hot vector with the value set outside of the training range to 2 (e.g.~[2, 0, 0, 0, 0, 0, 0, 0, 0, 0]), while keeping the uniform latent variables ($z$) constant across the 10 groups. 1000 outputs were thus annotated. 

Similarly to the 5-word model in Section \ref{8011}, the generated data suggests that the Generator learns to associate each lexical item with a unique representation. To test the significance of the latent code as a predictor, the coded annotated data were fit to a multinomial logistic regression model (as described in Section \ref{ciw5}).\footnote{\label{code10}The outputs were coded according to the following criteria: if transcription included ``su[ie][td]'', then \emph{suit}, if ``[\^{}s]e[ae]r'' then \emph{year}, if ``water'' then \emph{water}, if ``dar'' then \emph{dark}, if ``greas'' then \emph{greasy}, if ``[kc].*r'' then \emph{carry}, if ``[ao][wia]ly'' then \emph{oily}, if ``rag'' then \emph{rag}, if ``as'' then \emph{ask}, if ``li'' then like.} The AIC test suggests that the latent code is a significant predictor ($AIC=4555.6$ for an empty model vs.~1909.4 for a model with the predictor). 

\begin{figure}
\centering
\includegraphics[width=.5\textwidth]{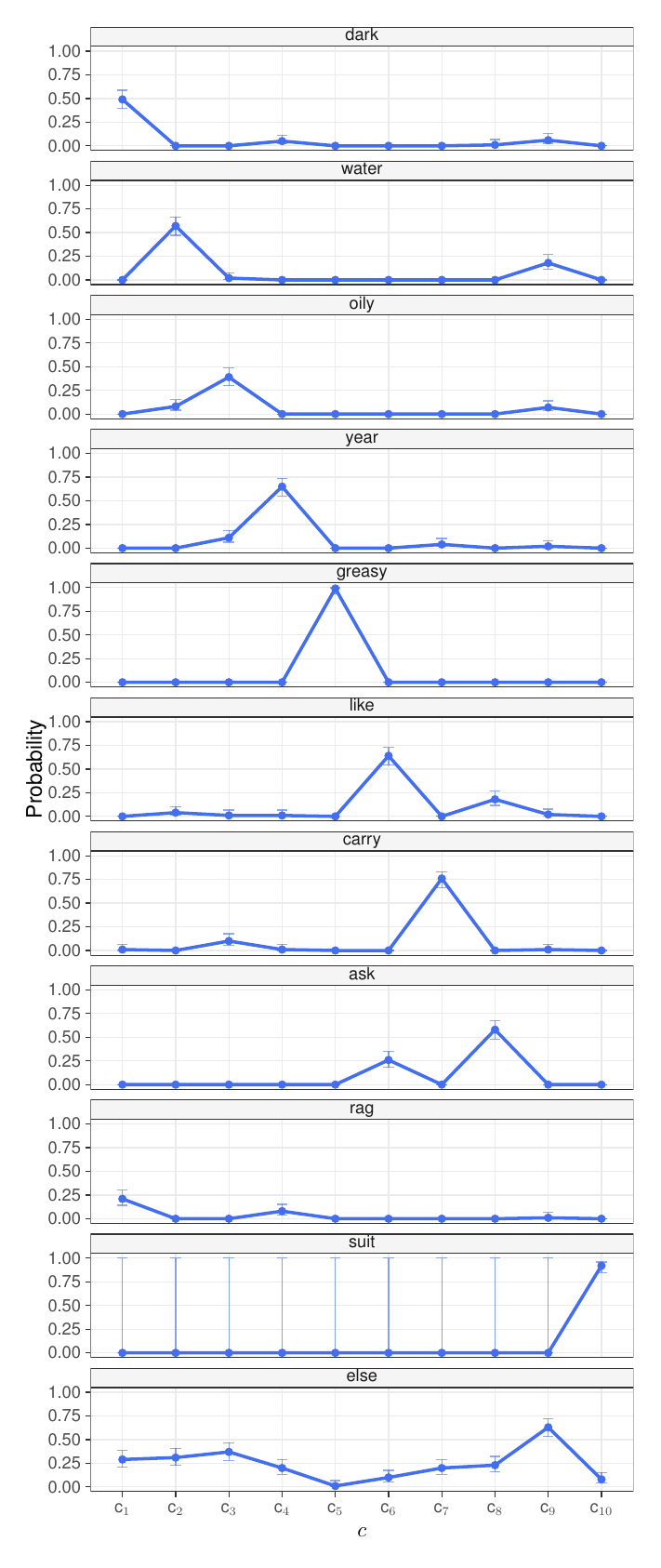}

\caption{\label{ceffect10}Estimates of a multinomial logistic regression model with coded transcribed outputs as the dependent variable and the latent code with ten levels that correspond to the ten unique one-hot vectors in the model  trained on ten lexical items from TIMIT after 27103 steps. }
\end{figure}

Estimates from the multinomial logistic regression model in Figure \ref{ceffect10} illustrate that each unique one-hot vector is associated with a unique lexical item. Each lexical item has a single substantial peak in estimates per latent code. The only exception appears to be \emph{rag} without a clear representation. The highest proportion of \emph{rag} appears for $c_1  =2$ at approximately 20\%. However, this particular latent code ($c_1  =2$) already outputs a substantially higher proportion of \emph{dark}. It thus appears that the Generator fails to generate outputs such that the difference between the two outputs would be substantial. There is a high degree of phonetic similarity precisely between these two lexical items: the vowels [\textipa{\ae}] and [\textipa{A}] are acoustically similar and both lexical items contain a rhotic [\textipa{\*r}] and a voiced stop. Success rates for the other nine lexical items range from 39\%--99\% in raw counts (for all raw counts, see the Appendix). 

\begin{figure}
\centering
\includegraphics[width=1\textwidth]{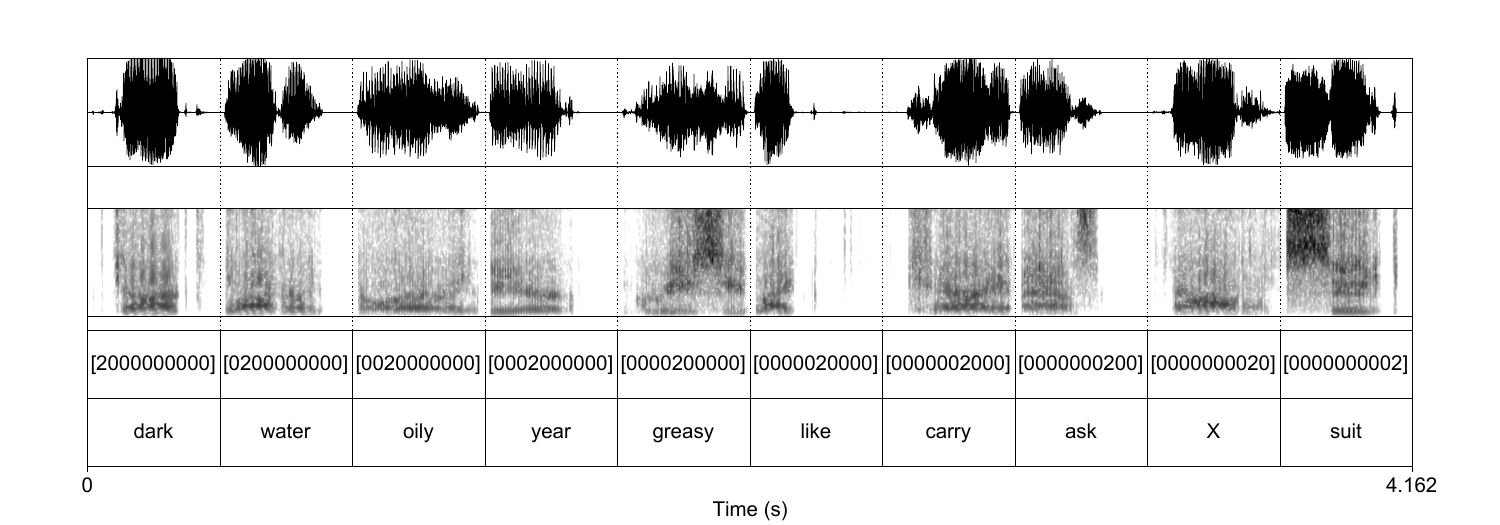}

\caption{\label{darkyear}Waveforms and spectrograms (0-8000 Hz) of generated outputs (of a model trained on 10 items after 27103 steps) when only the latent code is manipulated and the remaining 90 latent random variables are kept constant across all 10 outputs. Transcriptions (by the author) suggest that each lexical item is associated with a unique representation.}
\end{figure}

To illustrate that the network learns to associate lexical items with unique values in the latent code (one-hot vector), we generate outputs by manipulating the one-hot vector for each value and by keeping the rest of the latent space ($z$) constant. Such a manipulation can result in generated samples, where each latent space outputs a distinct lexical item associated with that value ([20000000000] outputs \emph{dark}, [02000000000] \emph{water}, etc.).\footnote{Often each series outputs one or two divergences from the ideal output.} Note that the acoustic contents of the generated outputs that correspond to each lexical item are substantially different (as illustrated by the spectrograms in Figure \ref{darkyear}), which means that the latent code ($c$) needs to be strongly associated with the individual lexical items, given that all the other 90 variables in the latent space (the $z$-variables which constitute 90\% of all latent space)   are kept constant and that the entire change of the output occurs only due to change of the latent code $c$.  In other words, by only changing the latent code and setting the variables to desired values while keeping the rest of the latent space constant, we can generate desired lexical items with the Generator network.

\subsection{fiwGAN on 8 lexical items}
\label{fiwGANon8}

To evaluate lexical learning in a fiwGAN architecture, we train the fiwGAN model with three featural variables ($\phi$). Because the latent code in fiwGAN is binomially distributed, three featural variables correspond to $2^3=8$ categories.   The model was trained on  8 content lexical items with more than 600 attestations in the TIMIT database (listed in Table \ref{8items}).  The model used for the analysis was trained after 20026 steps which correspond to a similar number of epochs as the 10-word ciwGAN model in Section \ref{ciw10} ($\sim$ 1241 epochs). Like for the ciwGAN models (Sections \ref{ciw5} and \ref{ciw10}), we generate 100 outputs for  each unique binary code given the 3 featural variables with the values of the features set outside of the training range to 2 instead of 1: [0, 0, 0], [0, 0, 2], [0, 2, 0], [0, 2, 2], [2, 0, 0], etc. 

\begin{table}\centering
\begin{tabular}{llc}\hline\hline
word&IPA&data points\\
\hline
ask&[\textipa{"\ae sk}]&633\\
carry&[\textipa{"k\super h\ae\*ri}]&632\\
dark&[\textipa{"dA\*rk}]&644\\
greasy&[\textipa{"g\*risi}]&630\\
like&[\textipa{"laIk}]&697\\
suit &\textipa{["sut]}&630\\
water& \textipa{["wOR\textrhookschwa]}&649\\
year &\textipa{["jI\*r]}&650\\\hline
\textbf{Total}&&5165\\
\hline\hline
\end{tabular}
\caption{\label{8items}Eight content lexical items from the TIMIT database used for training in the fiwGAN architecture (Section \ref{fiwGANon8}) with their corresponding IPA transcription (based on general American English) and counts of data points for each item. }

\end{table}

As expected, learning in the fiwGAN architecture is more challenging compared to ciwGAN. The network has only $\log_2(n)$ variables to encode $n$ lexical items (compared to $n$ variables for $n$ classes in ciwGAN). Despite the latent space for lexical  learning being highly reduced, an analysis of generated data in the fiwGAN architecture suggests that the Generator learns to associate each binary code with a distinct lexical item (for an additional test, see Section \ref{up}). 

 To test significance of the featural code ($\phi$) as a predictor, the annotated data were fit to a multinomial logistic regression model as in Section \ref{ciw5} and \ref{ciw10}. The dependent variables are again coded transcriptions\footnote{\label{code8}The outputs are coded as described in fn.~\ref{code10} for the 10-word ciwGAN model, except that if ``[ae].*[sf]'', then \emph{ask}, because outputs contain a large proportion of \emph{s}-like frication noise that can also be transcribed with \emph{f}.} and the independent variable is the featural code ($\phi$) with the eight unique levels as predictors: each for unique binary code. The difference in AIC between the model that includes the unique featural codes as predictors ($\phi$) and the empty model (2038.5 vs.~3409.7) suggest that featural values are significant predictors. 
 
\begin{figure}
\centering
\includegraphics[width=.5\textwidth]{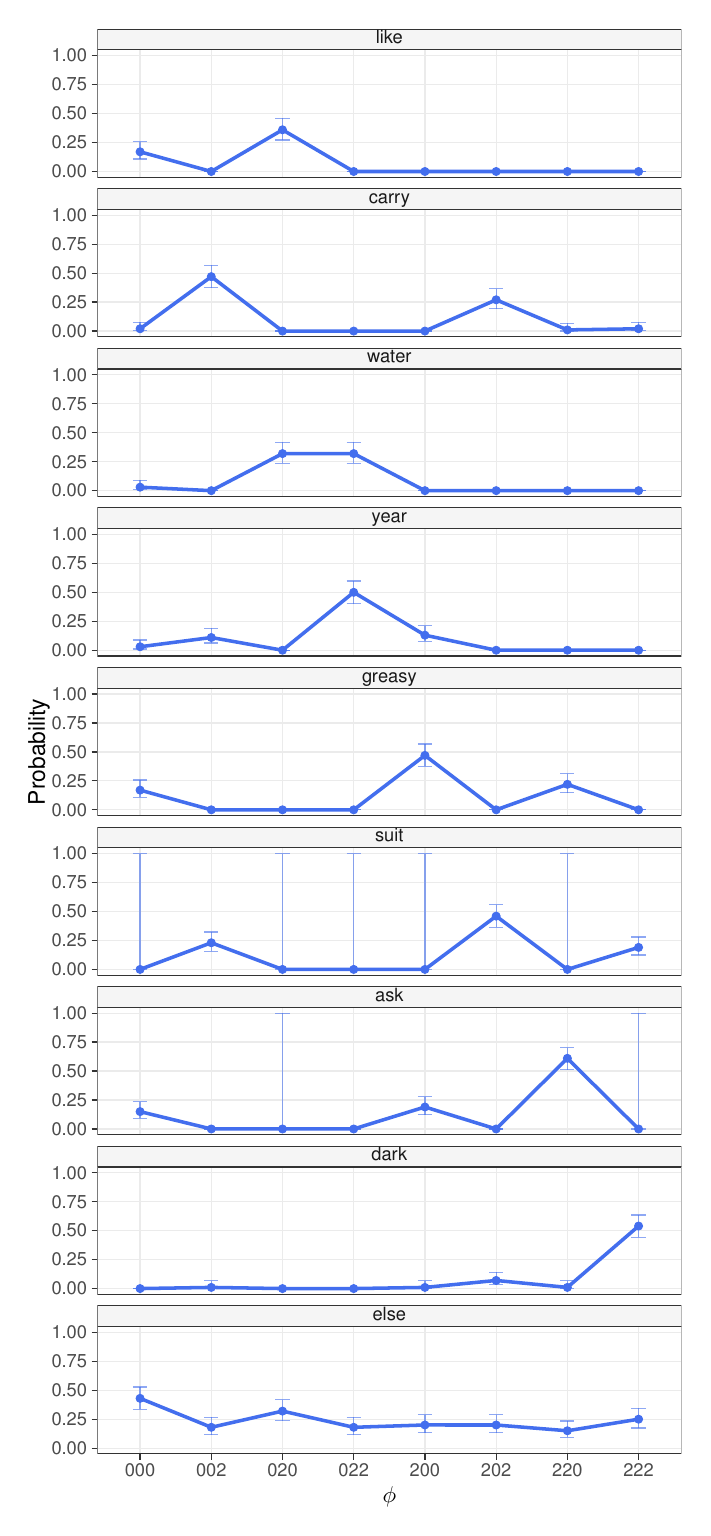}

\caption{\label{ceffect8}Estimates of a multinomial logistic regression model with coded transcribed outputs as the dependent variable and the latent code with eight levels that correspond to the eight unique binary codes in the model  trained on eight lexical items from TIMIT after 20026 steps. }
\end{figure}

 Estimates of the regression model in Figure \ref{ceffect8} illustrate that most lexical items receive a unique featural representation. Six out of eight lexical items (\emph{dark, ask, suit, greasy, year} and \emph{carry}) all have distinct latent featural representations that can be associated with these lexical items. Success rates for the six items have a mean of 50.8\% (in raw counts) with the  range of 46\% to 61\%. Crucially, there appears to be a single peak in regression estimates per lexical item for these six words, although the peaks are less prominent compared to the ciwGAN architecture (expectedly so, since learning is significantly more challenging in the featural condition). \emph{Water} and \emph{like} are more problematic: [0, 2, 0] outputs \emph{like} and \emph{water} at approximately the same rate. It is possible that learning of the two lexical items is unsuccessful. Another possibility is that  [0, 2, 0] is the underlying representation of \emph{water} because it is \emph{water}'s most frequent code that is not already taken by another lexical item. According to the guidelines in Section \ref{ciw5}, \emph{like} would have to be represented by [0, 0, 0], because it outputs the highest proportion of \emph{like} that is not already taken for another lexical items. That this assignment of underlying values of each featural representation is valid is additionally suggested by another test in Section \ref{up}.

In the fiwGAN architecture, we can also test significance of each of the three unique features ($\phi_1$, $\phi_2$, and $\phi_3$). The annotated data were fit to the same multinomial logistic regression model as above, but with three independent variables: the three features each with two levels (0 and 2). AIC is lowest when all three variables are present in the model (2135.5) compared to when $\phi_1$, $\phi_2$, or $\phi_3$ are removed from the model (2527.2, 2413.0, and 2773.3, respectively).

\subsubsection{Featural learning}
\label{fl}

An advantage of the fiwGAN architecture is that it can model classification (i.e.~lexical learning) and featural learning of phonetic and phonological representations simultaneously. We can assume that lexical learning is represented by the unique binary code for each lexical item. Phonetic and phonological information can be simultaneously encoded with each unique feature ($\phi$). That phonetic and phonological information is learned with binary features has been the prevalent assumption in linguistics for decades \citep{clements85,hayes09}. Recently,  neuroimaging evidence suggesting that phonetic and phonological information is stored as signals that approximate  phonological features has been presented in \cite{mesgarani14}.

An analysis of featural learning in fiwGAN --- how featural codes simultaneously represent unique lexical items and phonetic/phonological representations, can be performed by using logistic regression as proposed in the following paragraphs as well as with a number of exploratory techniques described in this section.

 Three out of eight lexical items used in training of the fiwGAN model include the segment [s]: a voiceless alveolar fricative with a distinct phonetic marker --- a period of frication noise. The assumed binary codes for the three items containing [s], \emph{ask}, \emph{greasy}, and \emph{suit} are [2, 2, 0], [2, 0, 0], and [2, 0, 2] (see Figure \ref{ceffect8}). We observe that value 2 for feature $\phi_1$ is common to all three of the lexical items containing an [s].

\begin{figure}
\centering
\includegraphics[width=.45\textwidth]{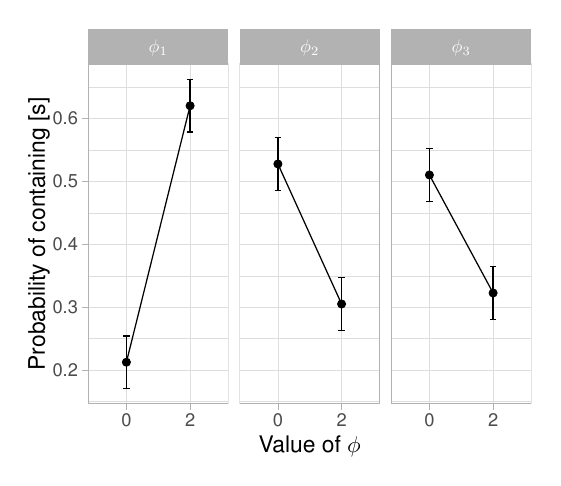}

\caption{\label{featurestand}Fitted values with 95\% CIs of a logistic regression model with presence of [s]-like frication in the transcribed outputs as the dependent variable and the three features $\phi_1$, $\phi_2$, and $\phi_3$ as predictors.}
\end{figure}

To test the effects of $\phi_1$ on presence of [s] in the output, 800 annotated outputs (100 for each of the eight unique binary codes) were fit to a logistic regression model. The dependent variable is presence of a \emph{s}-like frication noise: if a transcribed output contains an \emph{s}, \emph{z}, or \emph{f}, the output is coded as success. The independent predictors in the model are the three features without interactions: $\phi_1$, $\phi_2$, and $\phi_3$, each with two levels (0 and 2). Figure \ref{featurestand} features estimates of the regression model. While all three features are significant predictors, the effect appears to be most prominent  for $\phi_1$. 

It is possible that the Generator network in the fiwGAN architecture uses feature $\phi_1$ to encode presence of segment [s] in the output. This distribution can also be due to chance. Further work is needed to test whether presence of phonetic/phonological elements in the output can be encoded with individual features. Two facts from the generated data, however,  suggest that the Generator in the fiwGAN architecture associates $\phi_1$ with presence of [s]. 

\begin{figure}
\centering
\includegraphics[width=.55\textwidth]{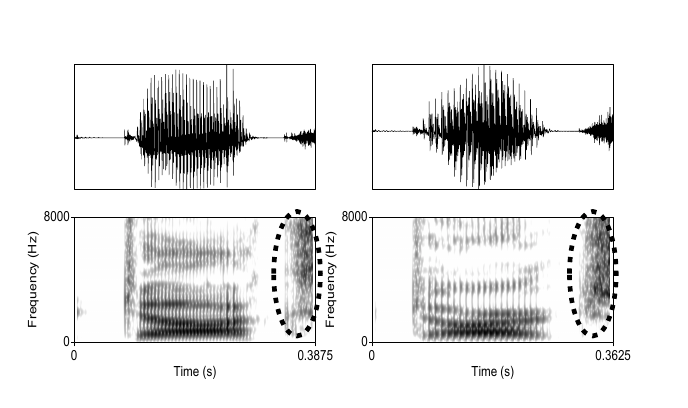}

\caption{\label{darks}Waveforms and spectrograms (0-8000 Hz) of two lexical items \emph{dark} from TIMIT with a clear \emph{s}-like frication noise during the aspiration after the closure of [k] (highlighted). }
\end{figure}

First, while \emph{ask}, \emph{greasy}, and \emph{suit} all have $\phi_1=2$ in common, the fourth unique featural code with $\phi_1=2$ ([2, 2, 2]) is associated with \emph{dark}. Spectral analysis of lexical item \emph{dark} in the training data reveals that aspiration of [k] in \emph{dark}  is in the training data from TIMIT frequently realized precisely as an alveolar fricative [s] (likely due to contextual influences).\footnote{In many TIMIT sentences, \emph{dark} appears before \emph{suit} which causes the aspiration of [k] to be influenced by the following [s].} Approximately 27\% data points for \emph{dark} in the training data from TIMIT contain a [s]-like frication noise during the aspiration period of [k].\footnote{This estimate is based on acoustic analysis of the first 100 training data points from the TIMIT database.} Figure \ref{darks} gives two such examples from TIMIT of \emph{dark} with a clear frication noise characteristic of an [s] sound after the aspiration noise of [k].  In other words, 3 lexical items in the training data contain an [s] as part of their phonemic representation and therefore feature it consistently. The Generator outputs data such that a single feature ($\phi_1=2$) is common to all three items. An additional item often involves a \emph{s}-like element and the network uses the same value ($\phi_1=2$) for its unique code ([2, 2, 2]). There is approximately a 8.6\% chance this distribution is random (of 70 possible featural code assignment for eight items, four of which contain some phonetic feature such as [s], six or 8.6\% combinations contain the same value in one feature). 

 As already mentioned, the network outputs mostly \emph{dark} for the featural code [2, 2, 2], but a substantial portion of outputs also deviate from \emph{dark}. A closer look at the structure of the innovative outputs for the [2, 2, 2] code reveals that a substantial proportion of them (35) contain an [s]. As a comparison, other unique codes with $\phi_1$ set at the opposite value 0 ([0, 0, 0], [0, 0, 2], [0, 2, 0], [0, 2, 2]) output 43, 41, 1, and 0 outputs containing an \emph{s}, \emph{z}, or \emph{f}. In other words, for two unique codes given $\phi_1=0$, the network generates 0 or 1 output containing an \emph{s}-like segment. For the two other codes, the network generates outputs with a similar rate of \emph{s}-containing sequences as [2, 2, 2] (\emph{dark}).  However, the motivation for an \emph{s-}containing output in [0, 2, 2] is clear: \emph{year} is in three training data points actually realized as [\textipa{SI\*r}] (\emph{shear}). The [0, 0, 0] does not have a distinct underlying lexical item, so the high proportion of outputs with [s] is not unexpected. 

\begin{figure}
\centering
\includegraphics[width=.8\textwidth]{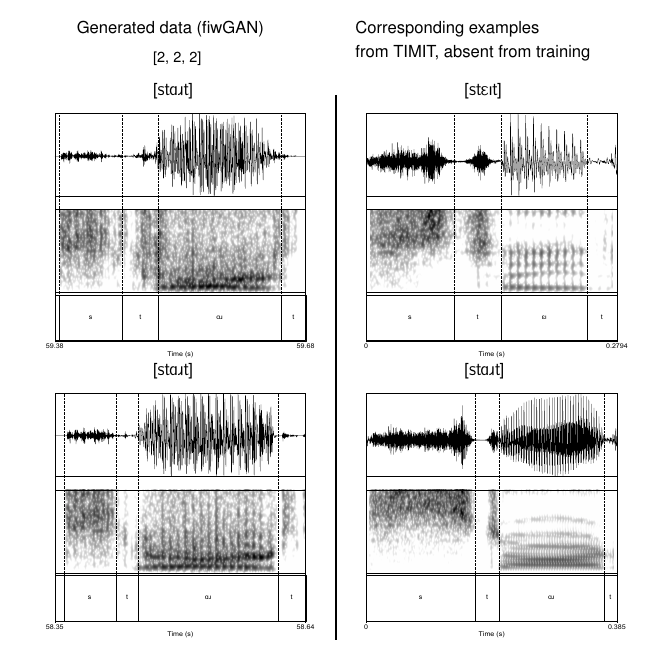}

\caption{\label{start}(left) Waveforms and spectrograms (0-8000 Hz) of two innovative outputs for $\phi=[2, 2, 2]$ transcribed as [\textipa{"stA\*rt}]. The fiwGAN network trained after 20026 steps thus outputs innovative sequence [st] that is absent from the training data, but is a linguistically interpretable output that can be interpreted as adding [s] to \emph{dark}.    (right) Waveforms and spectrograms (0-8000 Hz) of two lexical items from TIMIT that are \emph{not} part of training data, but illustrate that the innovative [st] sequence in the generated data is acoustically very similar to the [st] sequence in human outputs that the network has no access to. }
\end{figure}

\begin{figure}
\centering
\includegraphics[width=.6\textwidth]{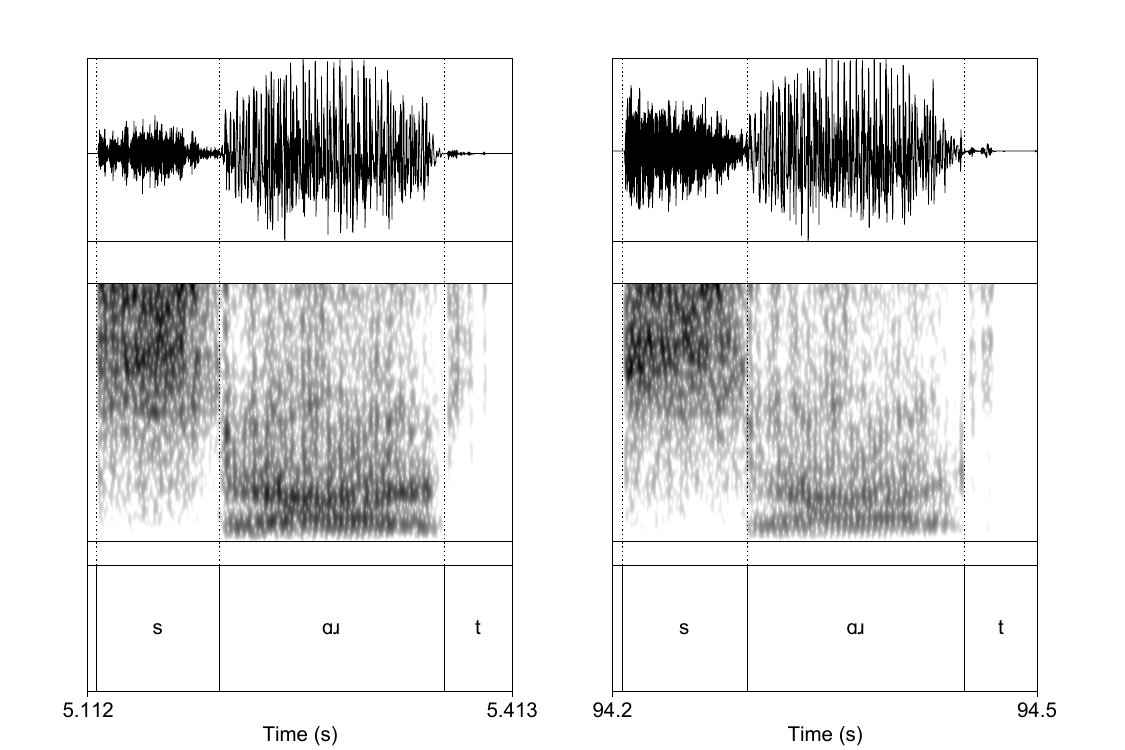}

\caption{\label{sart}Waveforms and spectrograms (0-8000 Hz) of two innovative outputs for $\phi=[2, 2, 2]$ transcribed as [\textipa{"sA\*rt}].}
\end{figure}

The second piece of evidence suggesting that ($\phi_1=2$) represents presence of [s] in the output are innovative outputs that violate the training data distribution. The majority of \emph{s}-containing outputs when $\phi_1=0$ are non-innovative sequences that correspond to  lexical items from the training data. The most notable feature of the \emph{s}-containing outputs for [2, 2, 2] (\emph{dark}), on the other hand, is their innovative nature. Sometimes, these outputs can indeed be transcribed as \emph{suit}, but in some cases the Generator outputs an innovative sequence that violates training data but is still linguistically interpretable. In fact, some of the outputs with [2, 2, 2] are directly interpretable as adding an [s] to the underlying form \emph{dark}. Two innovative sequences that can be reliably transcribed as  \emph{start} [\textipa{"stA\*rt}] are given in Figure \ref{start} and additional two transcribed as \emph{sart} [\textipa{"sA\*rt}] in Figure \ref{sart}.  The network is never trained on a [st] sequence, let alone on the lexical item \emph{start}, yet the innovative output is linguistically interpretable and remarkably similar to  the [st] sequence in human outputs that the network never ``sees''. Spectral analysis illustrates a clear period of frication noise characteristic of [s] followed by a period of silence and release burst characteristic of a stop [t]. Figure \ref{start} provides two examples from the TIMIT database with the [st] sequence that was never part of the training data, yet illustrates how acoustically  similar the innovative generated outputs in the fiwGAN architecture are to real speech data. This example constitutes one of the prime cases of high productivity of deep neural networks (for a recent survey on productivity in deep learning, see \citealt{baroni19}). 

\subsubsection{Underlying representations}
\label{up}

\cite{begus19} argues that the underlying value of a feature can be uncovered by manipulating a given feature well beyond the training range. For example, \cite{begus19} proposes a technique for identifying variables that correspond to phonetic/phonological properties in the outputs. By setting the values of the identified features well beyond the training range, the network almost exclusively outputs the desired property (e.g.~a segment [s] in the output). 

This effect of the underlying value of a variable is even more prominent in the fiwGAN architecture. When the values of latent features ($\phi$) are set at 0 and 2, success rates appear at approximately 50\% (Figure \ref{ceffect8}). Value 2 was chosen for analysis in Section \ref{fiwGANon8}, because non-categorical outcomes yield more insights into learning. However, we can reach almost categorical accuracy when the values are set substantially higher than 2. For example, when we generate outputs with values of featural code set at 5 ([5, 5, 5]), the network generates 93/100 outputs\footnote{Counts in this sections are performed by the author only.} that can be reliably transcribed as \emph{dark} and another 7 that closely resemble \emph{dark}, but have a period of frication instead of the initial stop (\emph{sark}). With even higher values such as 15, the Generator outputs 100/100 samples transcribed as \emph{dark} for [15, 15, 15]. Similarly, [5, 5, 0] yields \emph{ask} in 97/100 cases; it yields an innovative output with final [i] in three cases. At [15, 15, 0], the Generator outputs 100/100 \emph{ask}. The success rates differ across featural codes, but value 15 triggers almost categorical outputs for most of them. [15, 0, 0] yields 93/100 \emph{greasy} (1 unclear and 6 \emph{ask}). For [15, 0, 15], the network outputs a [sVt] sequence for \emph{suit} (where V=vowel) 87/100 times. In 13 examples, the frication noise does not have a pronounced s-like frication noise, but is more distributed and closer to aspiration noise of [k]. The identity of the vowel is intriguing: the formant values are not characteristic of [u] (as in \emph{suit}), but rather of a lower more central vowel (F1 = 663 Hz, F2 = 1618 Hz, F3 = 2515 Hz for one listing). Since formant variability is high in the training data, the underlying prototypical representation likely defaults to a more central vowel. 

\begin{figure}
\centering
\includegraphics[width=.8\textwidth]{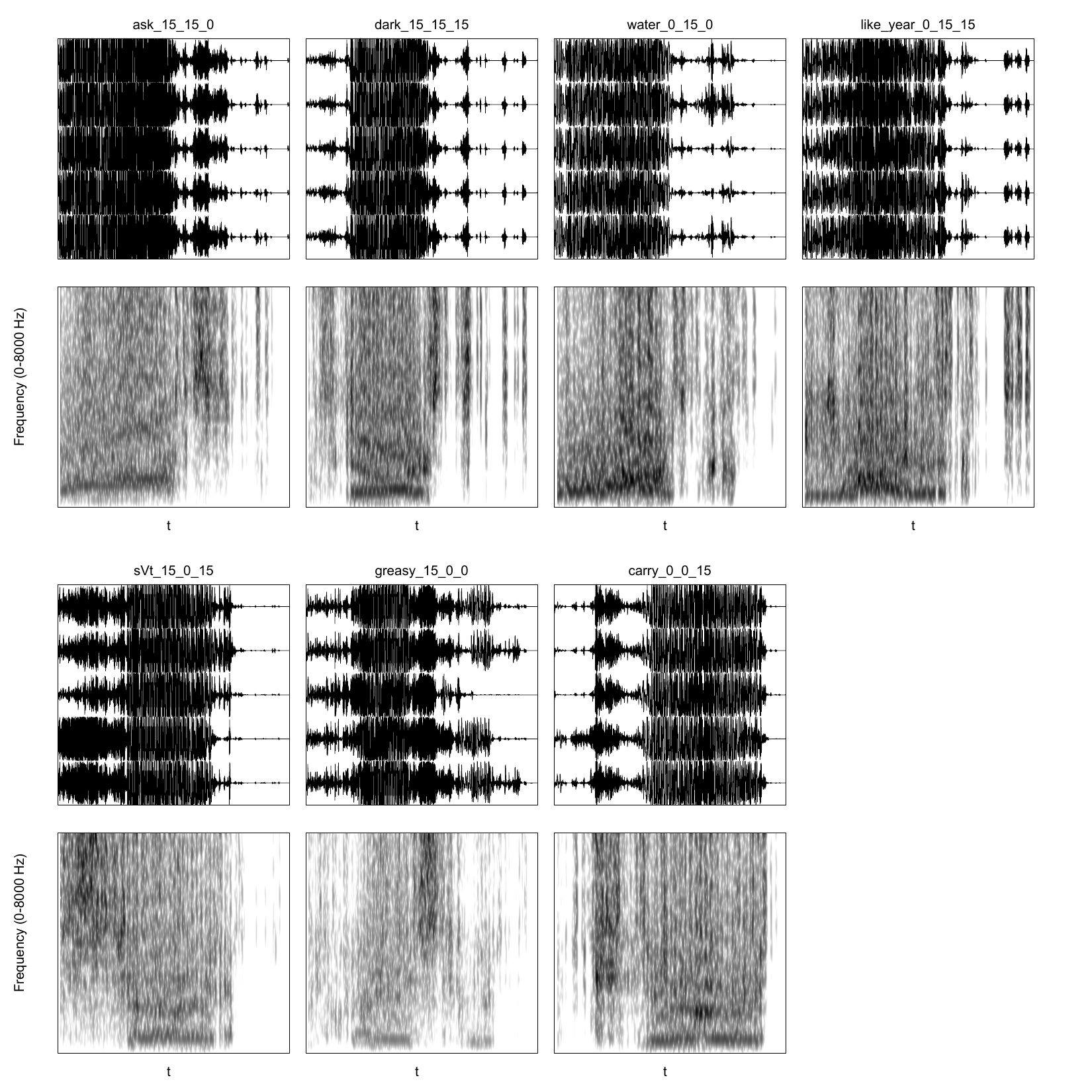}

\caption{\label{151515}Waveforms of the first five generated outputs for each featural code when values are set at 15. The waveforms clearly show that the outputs feature minimal variability. Below each waveform is a spectrogram (0--8000 Hz) of the first output (the topmost waveform). All seven outputs have the exact same values for 97 random latent variables ($z$); they only differ in the three featural codes $\phi$. }
\end{figure}

[0, 0, 15] yields \emph{carry} in  100/100 outputs, but in 13 of these outputs the aspiration noise of [k] is distributed with a peak in higher frequencies for an acoustic effect of [ts]. [0, 15, 0] yields 100/100 \emph{water}. The acoustic output is reduced to include only the main acoustic properties of \emph{water}: formant structure for [\textipa{wO}] followed by a consonantal period for [\textipa{R}] and a very brief (sometimes missing) vocalic period (Figure \ref{151515}).

The only two codes that do not yield straightforward underlying representations are [0, 0, 0] and  [0, 2, 2]. It appears that the Generator is  unable to strongly associate [0, 0, 0] with any lexical representation, likely due to lack of positive values in this particular code. This means that the network needs to learn underlying representations for two remaining lexical items with a single code: \emph{like} and \emph{year}, both likely associated with [0, 2, 2]. When set to [0, 5, 5], the Generator outputs both \emph{like} and \emph{year},\footnote{Occasionally, [0, 5, 5] also yields an output that can be characterized as \emph{water}.} but at [0, 10, 10] and [0, 15, 15] the underlying representation is an acoustic output that is difficult to characterize and is likely a blend of the two representations (acoustically closer to \emph{like}; see Figure \ref{151515}). Future analyses should thus include $\log_2(n)$ variables for $n-1$ classes in the fiwGAN architecture.

Another interesting fact emerges when we set the featural codes to high values such as 5 or 15. The outputs at these high values are minimally variable: the outputs are almost identical despite the  97 random latent variables $z$ being randomly sampled for each output, as illustrated by Figure \ref{151515}. It appears that the network associates unique featural codes with prototypical underlying representations of lexical items. When values are lower, other random latent variables ($z$) cause variation in the outputs, but the high values (such as 5 or 15) of the featural codes $\phi$ override this variation and reveal the  underlying lexical representation for each featural code.

The generative test with values set well above the training range strongly suggests that the Generator associates lexical items with unique codes and likely represents them with  prototypical acoustic values. The test also confirms that the assumed underlying lexical items identified with multinomial logistic regression (Figure \ref{ceffect8}) are correct.

\subsection{fiwGAN on the entire TIMIT}
\label{fiwGANonTimit}

To test how the proposed architecture scales to substantially larger training datasets, we train the fiwGAN architecture on the entire TIMIT database. TIMIT was sliced for words into the same format as described in Section \ref{8011}, which yields  54,378 lexical items in the training data. Because TIMIT contains approximately 6,229 unique lexical items, we train the fiwGAN network with 13 code variables, which allows for $2^{13} = 8,192$ unique classes.  The number of possible classes is  higher than the number of unique items (by about 1,963 or 24.0\%). This experiment thus also tests whether lexical learning emerges in a model in which the number of classes differs from the number of lexical items.  The network was trained for 179,528 steps ($\sim$1056 epochs).\footnote{The training on an NVIDIA GeForce GTX 1080 Ti takes approximately 160 steps per 300s for a total of $\sim$ 4 days and 12 hours. The models with higher number of steps face mode collapse issues, common in the GAN architecture.}

Training the networks to learn lexical items from the entire TIMIT  is a challenging task. First, neither the Generator (tested here for lexical learning) nor the Q-network (forcing the Generator to output informative data) have direct access to the data.  The only network that actually has a direct access to the training data is the Discriminator, which is the least relevant network in the architecture for learning lexical information. This stands in stark contrast with the autoencoder architecture, in which the network that learns lexical representations in its latent space has a direct access to the data.  Second, TIMIT is highly variable and contains lexical items that are phonetically highly reduced. Many items contain only one token per item, and in general, the token distribution of the lexical items varies substantially. The dimension of the vector representing lexical items in our models is highly reduced: only 13 binary code variables (13\% of the latent variables) are available to the model to encode identity of 6,229 unique items with 54,378 total tokens.

Based on inferential statistical tests in Sections \ref{ciw5}, \ref{ciw10}, and \ref{fiwGANon8}, we argued that the models show clear evidence for lexical learning. In a model with fewer codes, we can fit all annotations to a multinomial logistic regression model and perform hypothesis testing for lexical learning. In a larger model with the entire TIMIT as the training data, it is challenging to manually analyze outputs of all 8,192 binary codes given the thirteen latent features ($\phi$) in the model. Instead, we use the argument from the models with fewer variables and test lexical learning by generating data with a subset of possible latent codes. Evidence for lexical learning is evaluated in the following way: if the Generator outputs one lexical item more often than other lexical items for each unique code, it is reasonable to assume  (based on the results in Sections \ref{ciw5}, \ref{ciw10}, and \ref{fiwGANon8}) that the particular lexical item is the underlying learned representation of the corresponding  unique code, as is confirmed by statistical tests in  the  models with fewer variables. 

Despite the highly challenging learning task and the mismatch between the number of classes and the number of actual lexical items, the fiwGAN does appear to show evidence for lexical learning even when trained on the entire TIMIT dataset. We choose 70 out of possible 8,192 unique latent codes and generate sets of outputs, each for one of the 70 chosen codes. We only manipulate the latent code ($\phi$) across the 70 sets of outputs; all other 87/100 variables $z$ are identical  across the sets of outputs. The values of the latent code are set to 0 or either 1.0 or a slightly higher value of 1.1  (with no substantial differences in outputs observed among 1.0 or 1.1). As will be argued below, the Generator performs well on the lexical learning task even when the latent variables are not manipulated to marginal levels outside of the training range.\footnote{In fact, the Generator outputs unintelligible outputs when values are manipulated substantially outside of the training range (which stands in opposition to the models in Sections \ref{ciw5}, \ref{ciw10}, and \ref{fiwGANon8}). }

\begin{figure}
\centering
\includegraphics[width=.32\textwidth]{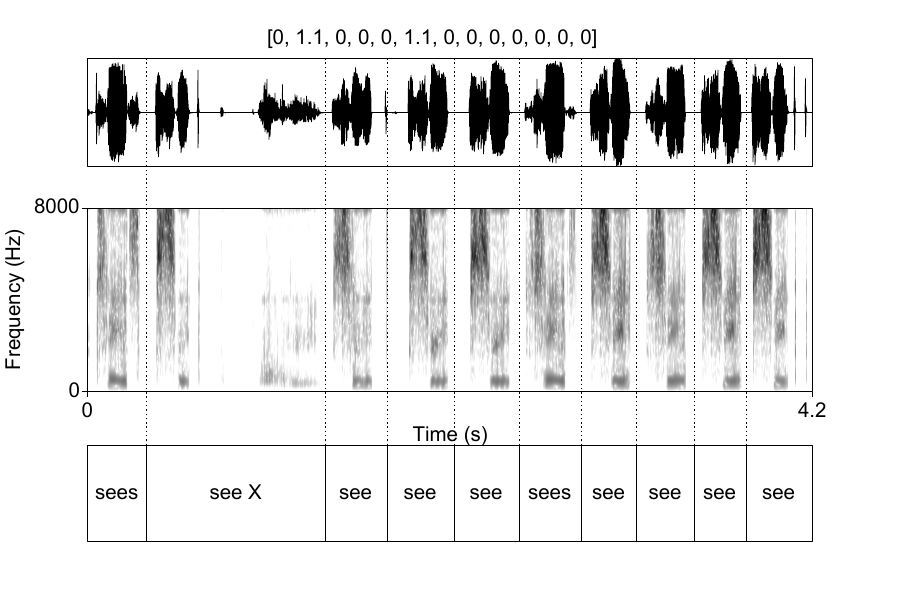}\includegraphics[width=.32\textwidth]{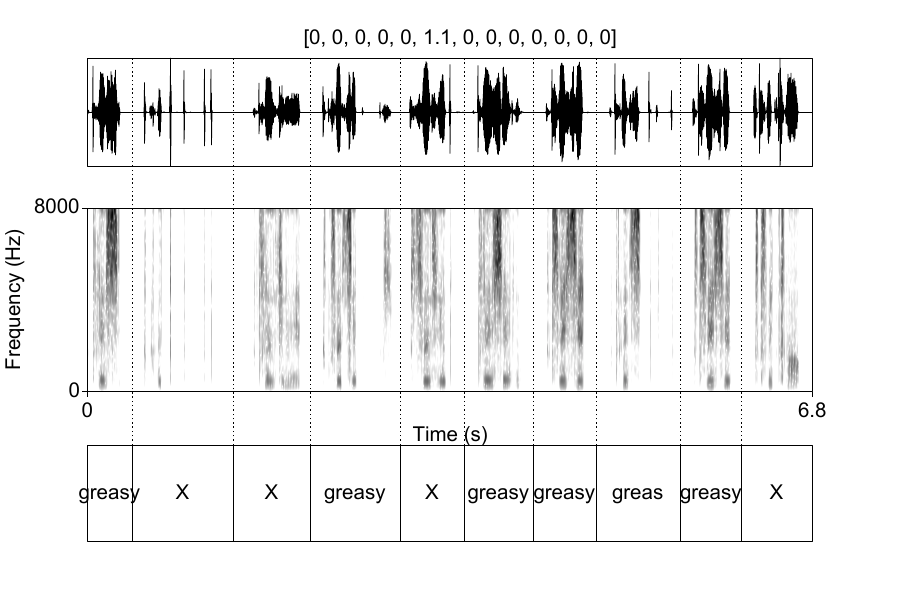}\includegraphics[width=.32\textwidth]{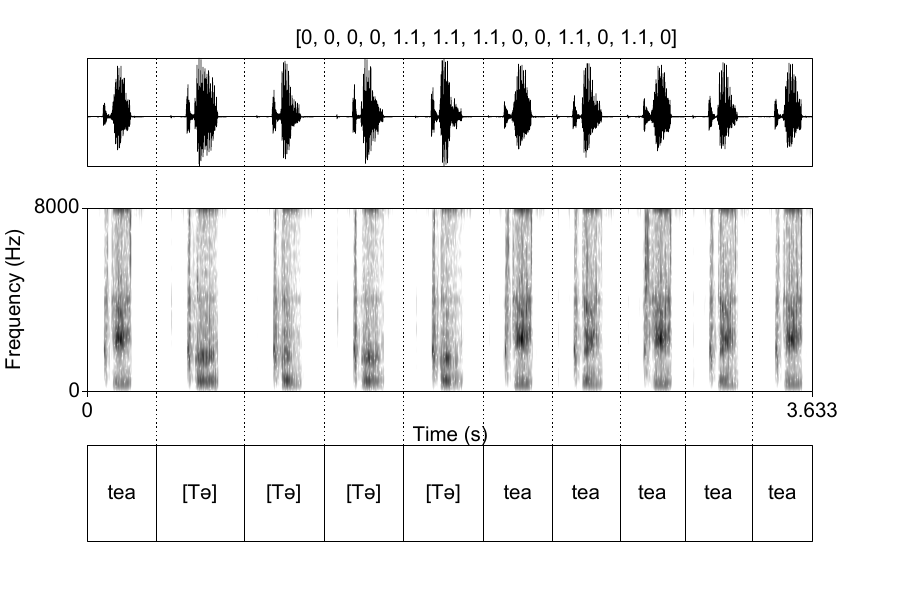}\\
\includegraphics[width=.32\textwidth]{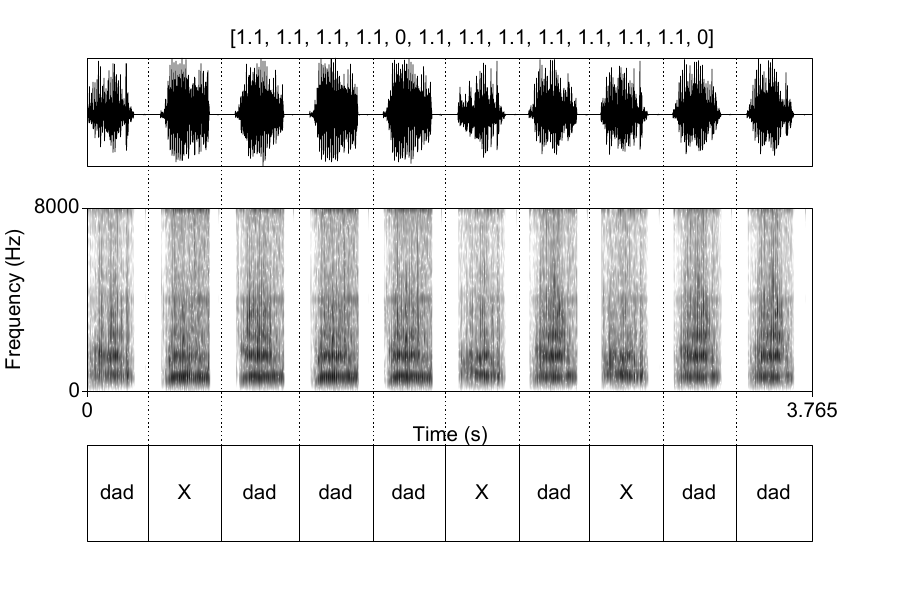}\includegraphics[width=.32\textwidth]{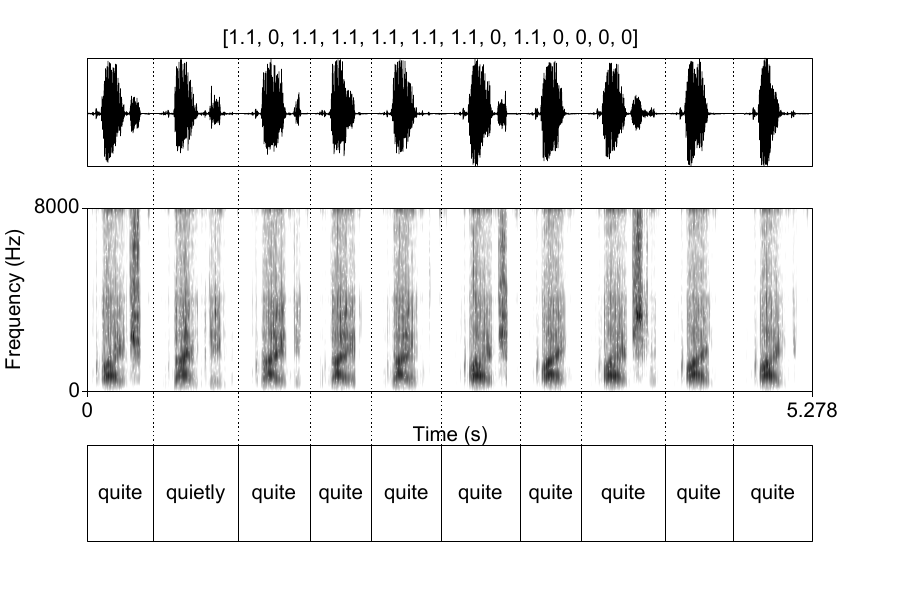}\includegraphics[width=.32\textwidth]{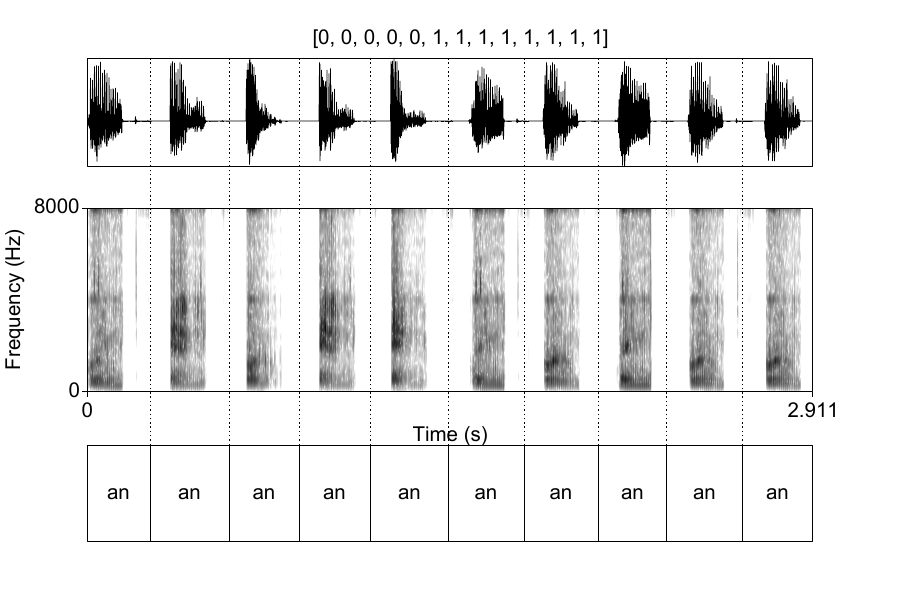}\\
\includegraphics[width=.32\textwidth]{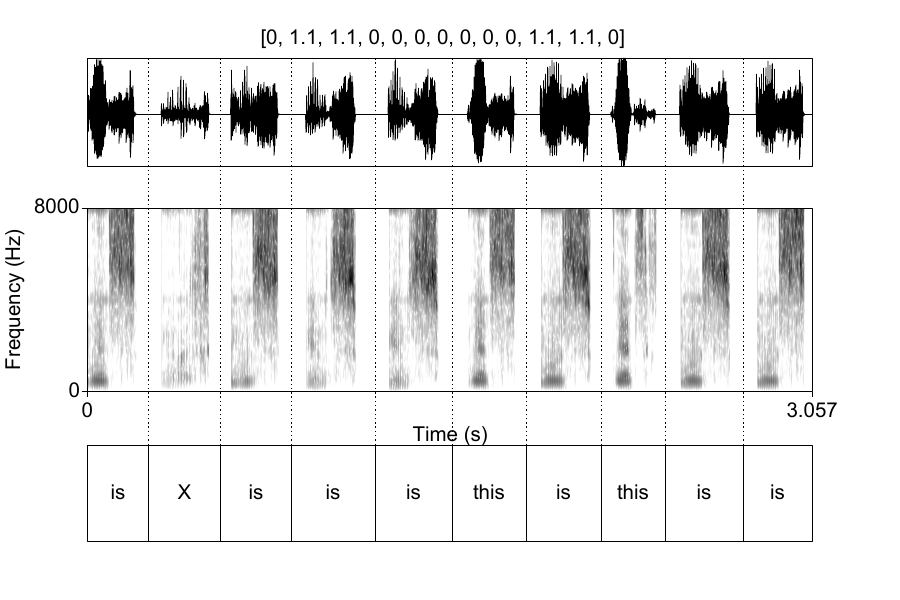}\includegraphics[width=.32\textwidth]{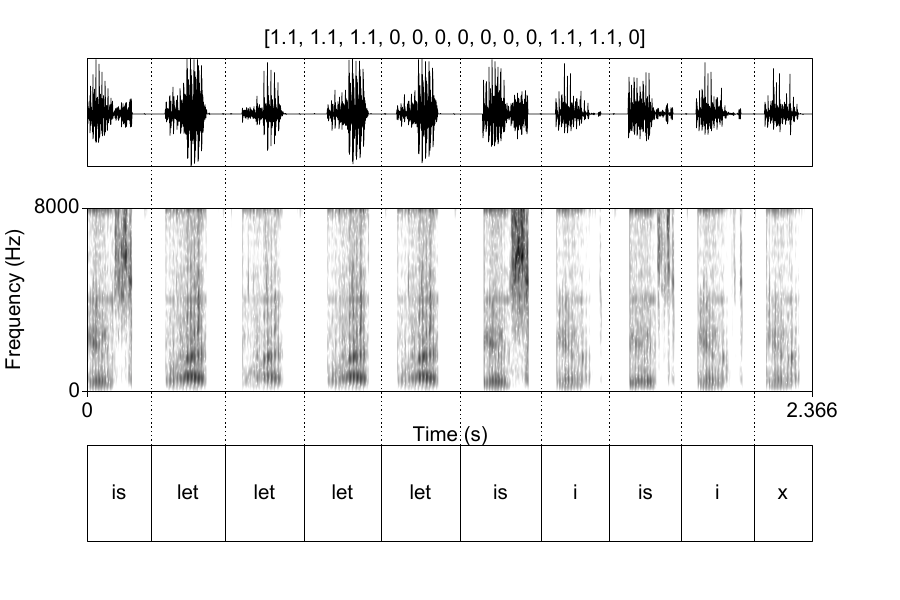}\includegraphics[width=.32\textwidth]{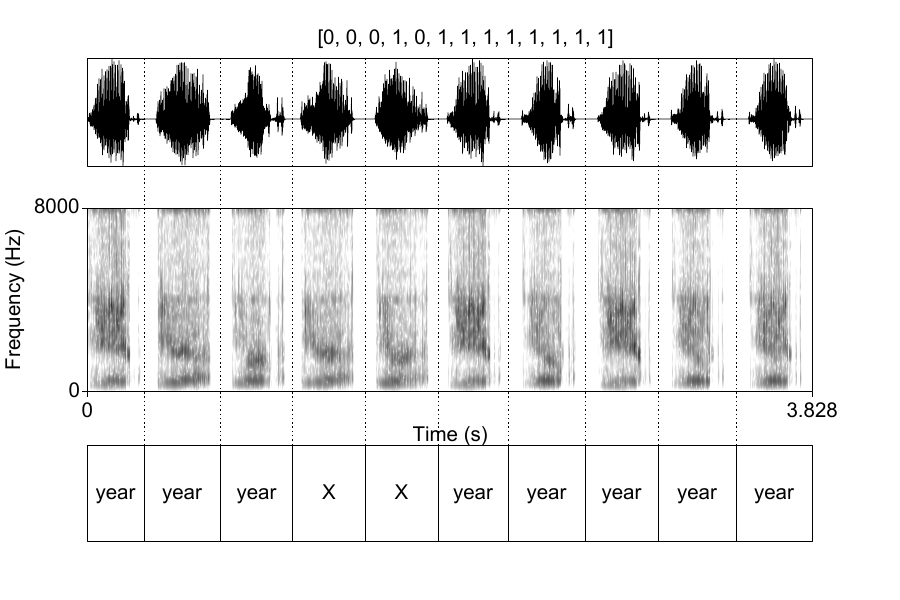}\\
\includegraphics[width=.32\textwidth]{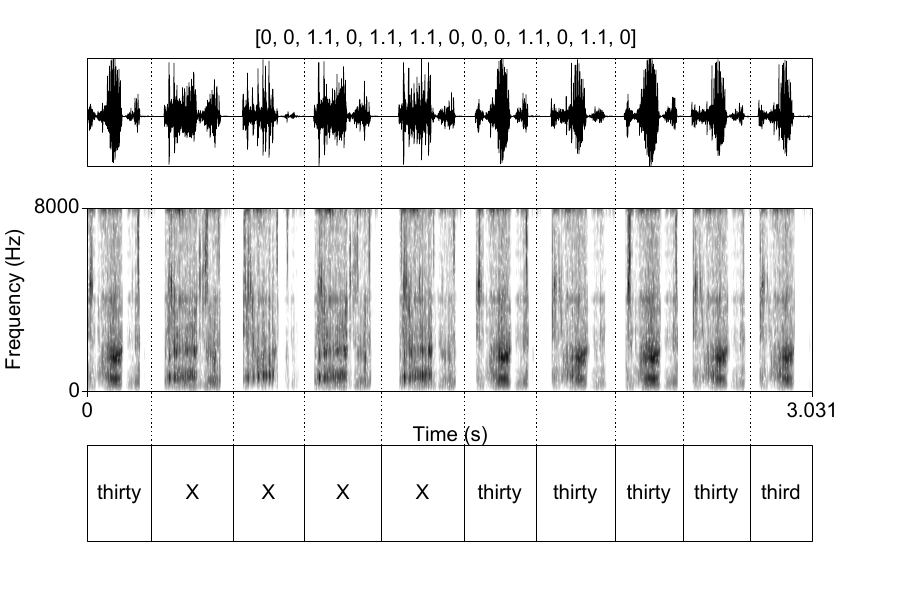}\includegraphics[width=.32\textwidth]{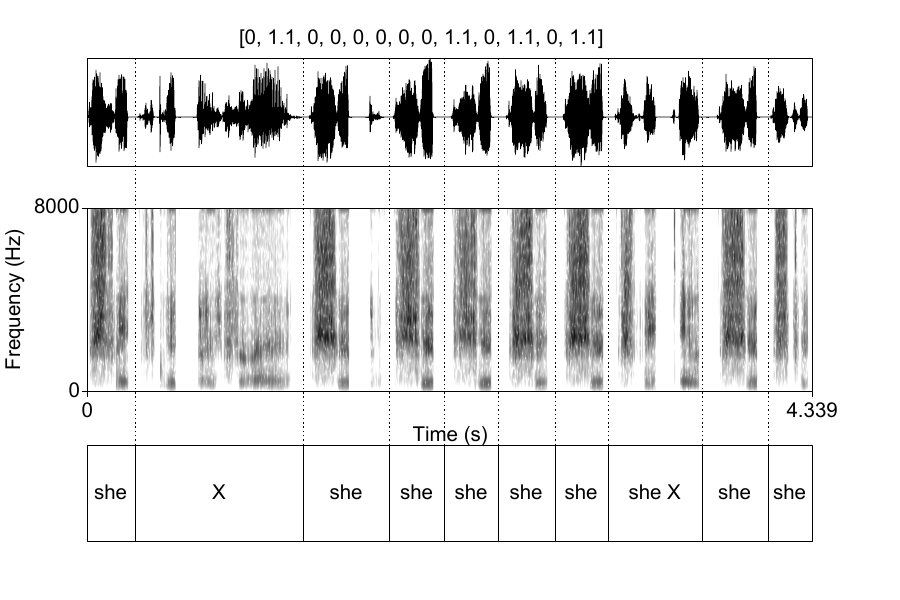}\includegraphics[width=.32\textwidth]{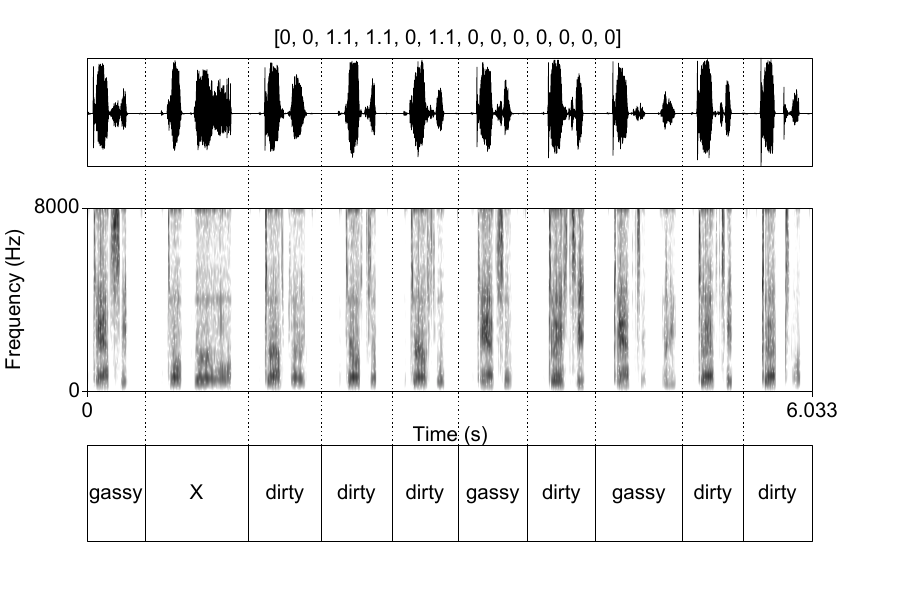}\\
\includegraphics[width=.32\textwidth]{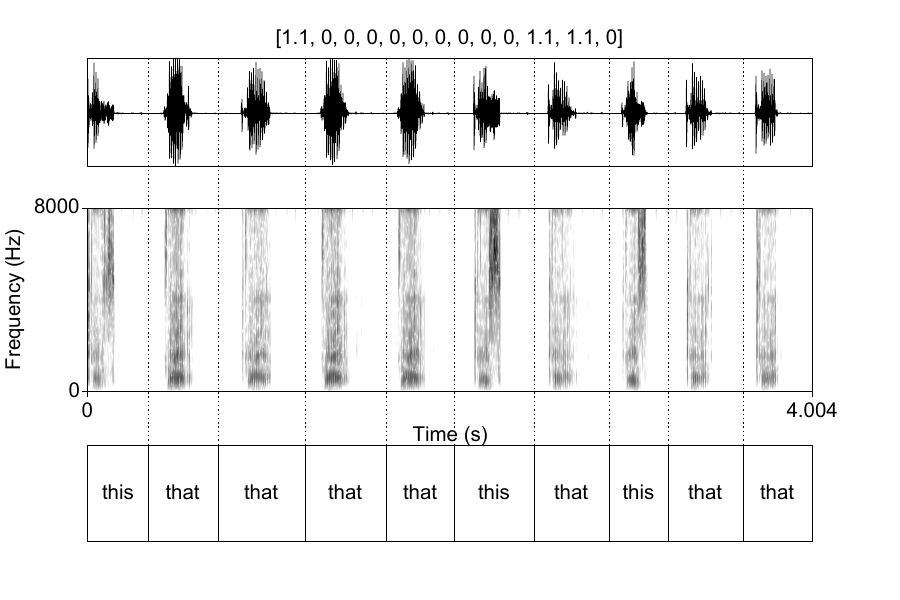}\includegraphics[width=.32\textwidth]{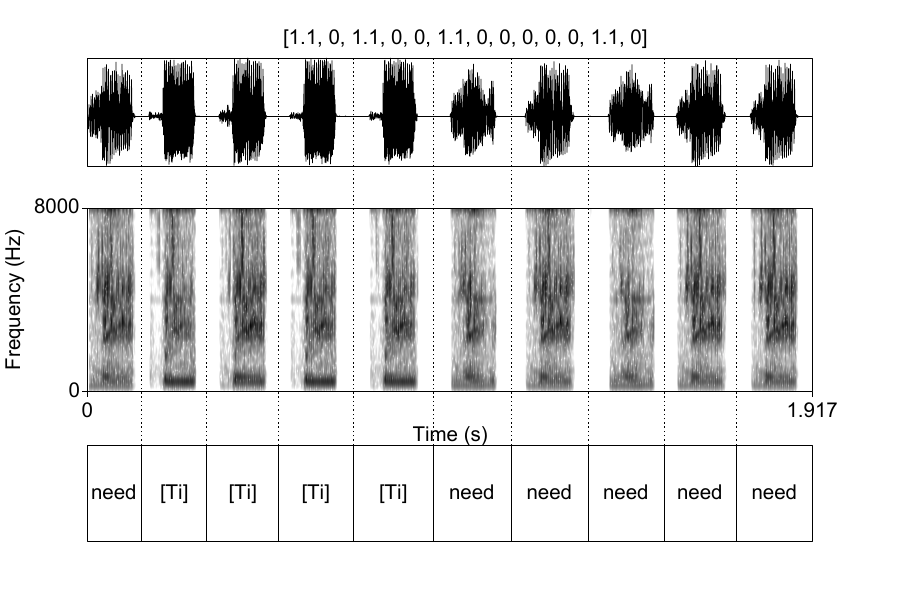}\includegraphics[width=.32\textwidth]{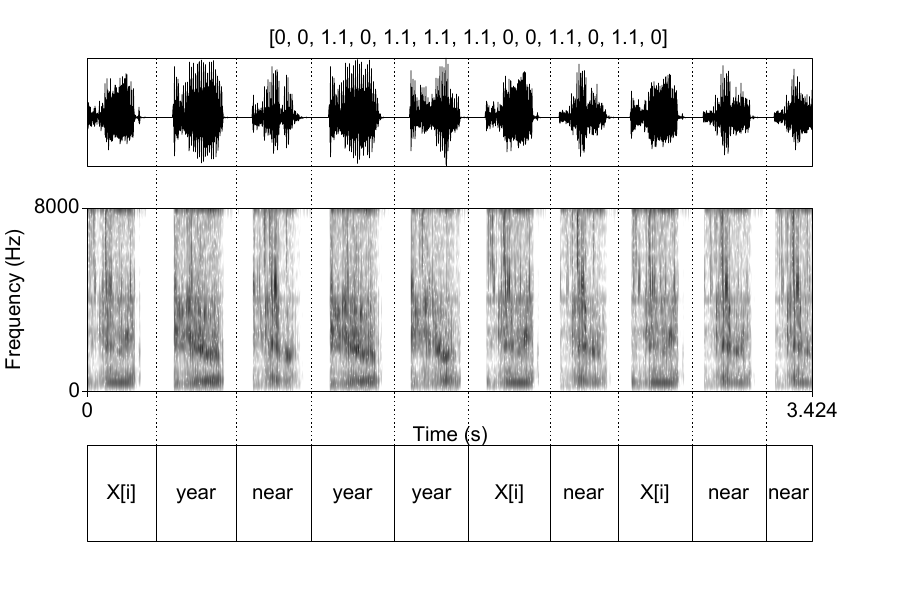}\\

\caption{Waveforms and spectrograms of fifteen sets of outputs, each for one unique latent code value. Each set contains ten outputs in which the other 87 latent variables $z$ are held constant across the fifteen sets. The Generator only outputs waveforms; spectrograms are given for the purpose of acoustic analysis. }
\label{13see}
\end{figure}

 Based on exploration of the 70 unique latent codes, three types of outcomes are observed: (i) the Generator outputs predominantly one clearly identifiable lexical item for a latent code; (ii) the Generator outputs predominantly one sequence of sounds for a latent code, but the actual underlying lexical item is difficult to establish; and (iii) the Generator outputs unclear and variable outputs for a latent code.\footnote{All outputs in the current section were analyzed and annotated by the author.} The first and desired outcome, where the Generator outputs almost exclusively one lexical item that is easily recognizable per one latent code, is the most frequent.\footnote{In some cases, the underlying lexical item is not the most frequent, but the only one that is identifiable.} For approximately 29 codes (out of 70 tested or 41.4\%), the Generator outputs mostly one clearly recognizable lexical item. Figure \ref{13see} shows the first ten generated outputs for fifteen out of 29 latent variables for which a clear underlying lexical item can be established. As already mentioned, the other latent variables ($z$) are identical in the ten outputs across the 15 sets. There is relatively little variation in the outputs: representation of a lexical item appears relatively uniform, as suggested by the spectrograms. 
 
It appears that the network learns both very frequent words (such as \emph{that} or \emph{is}), but also substantially less frequent words, such as \emph{let} which appears in  23 tokens or \emph{dirty} in only 15 tokens in the training data. A change of a single feature ($\phi$) can result in a substantial change in the output. For example, a change in $\phi_1$ from [0, 1.1, 1.1, 0, 0, 0, 0, 0, 0, 0, 1.1, 1.1, 0] to  [1.1, 1.1, 1.1, 0, 0, 0, 0, 0, 0, 0, 1.1, 1.1, 0] results from \emph{is} to \emph{let} (with occasional spill-over of \emph{is} as was also observed in the previous models). A change in $\phi_2$ from [0, 0, 0, 0, 0, 1.1, 0, 0, 0, 0, 0, 0, 0] to [0, 1.1, 0, 0, 0, 1.1, 0, 0, 0, 0, 0, 0, 0] results in \emph{greasy} and \emph{see}. Some words, on the other hand, can be represented with multiple latent codes and occasionally,  the change of one feature ($\phi$) does not result in a substantial change of the output. 

 While the networks appear to associate one lexical item per code even when there are more classes than lexical items (6,229 vs.~8,192), two or more unique codes can be associated with the same lexical item. There are approximately six such (mostly high frequency) words among the 70 explored unique latent codes: \emph{we}, \emph{the}, \emph{dirty}, \emph{that}, and \emph{let} are represented with two binary codes; \emph{is} with three codes. If the 70 latent codes are representative, we might expect a relative high number of the unique codes to be associated with a single lexical item. The proportion of unique codes actually associated with the same word (for each word associated with multiple codes), however, is probably substantially less than  2/70 (=234 of the total 8,192) or  3/70 (=350 of the total 8,192 in case of \emph{is}) precisely because the codes with which a single lexical item is associated more than once share many features. From the perspective of spoken term discovery, the multiple association is less problematic, but further work is needed to mitigate this problem. From a cognitive modeling perspective, the multiple association is not ideal. There might be two reasons for why this is also appealing. High frequency items have more variation in spoken language. In this way, the network can encode the same high frequency item, but with different phonetic properties. Second, the codes with which the network encodes the same item differ only in a single feature or two in five out of six cases (\emph{is} is the only item that differs in up to  six features). This allows for a unique representation of a lexical item in at least a large subset of variables.

Underlying values of the remaining 41 latent codes are not readily identifiable based on the generated outputs, but their outputs are not unstructured either. There is clear phonological structure in approximately 27 outputs (out of 70 or 38.6\%), but the exact lexical item is difficult to establish. Often, there is little variation in the output. For example, 8 (80\%) out of 10 generated outputs for a specific latent code feature a sequence of [s] and [p] and a following vowel ([spV] where V = vowel), but there is variation in the realization of the vowel and the exact lexical item cannot be identified. Similarly, one latent code outputs [\textipa{\*ris}V] in 7/10 outputs (especially frequent is  [\textipa{\*risi}]), yet it is difficult to establish which word  [\textipa{\*ris}V] represents.

There are two reasons for why the outputs in this group are not easily identifiable as lexical items.  Either the networks do represent unique lexical items with these codes, but the quality of the audio output is not high enough for a lexical item to be recognizable, or the networks encode combinations of phonemes with latent codes rather than unique lexical items. In other words, it is possible that sub-lexical units can be learned by the Generator and used to encode information. In the remaining 14 cases (20.0\%), the network generates an unclear output that is difficult to identify as a lexical item and the outputs do not show a uniform phonological structure either.

Some evidence for complex sublexical featural learning emerges in the fiwGAN trained on the entire TIMIT too. For example, if $\phi_{5-7}$ are set to 1.1 and $\phi_{12}$ is set to 0, the Generator frequently outputs a distinct initial velar stop [k] in six out of six selected sets of outputs with this structure of the latent code.\footnote{There are more outputs with this structure of features that have not been tested.} [k]-initial words are not reliably identified at the beginning of the word in any of the other 64 tested codes. While the items except \emph{quite} are not clear enough to fall in group (i), the initial [k] is clearly identifiable across the six sets. Apart from the initial [k], the outputs with this structure differ substantially in their phonetic properties (e.g.~outputs can approximate items such as \emph{culture}, \emph{car(t)}, \emph{quite}, \emph{color} or [kV]). It is reasonable to assume that the network encodes a sublexical phonetic property --- presence of initial [k] --- with a subset of latent features.

The results suggest that the Generator at least partially learns to encode lexical items with unique latent codes in the fiwGAN architecture even when trained on 54,378 tokens of 6,229 lexical items from TIMIT. Based on a subsample of latent codes,  the Generator outputs a clearly identifiable lexical item in  approximately 41.4\% of tested codes.  In most cases, the one lexical item is the majority output (as clear from Figure \ref{13see}) realized with relatively little phonetic variability. There are many properties of TIMIT data that the Generator could learn to encode into 8,192 classes, but it does appear that it is precisely lexical items that the network is learning. This experiment also suggests that the Generator learns to represent lexical items with unique codes even if the number of possible classes and the number of actual items do not match and even if there is a substantial difference in the frequency of individual items.

\section{Discussion and future directions}

This paper proposes two architectures for unsupervised modeling of lexical learning from raw acoustic data with deep neural networks. The ability to retrieve information from acoustic data in human speech is modeled with a Generator network that learns to output data that resemble speech and, simultaneously, learns to encode unique information in its outputs. We also propose techniques for probing how deep neural networks trained on speech data learn meaningful representations. 

The proposed fiwGAN and ciwGAN models are based on the Generative Adversarial Network architecture and its implementations in WaveGAN \citep{donahue19}, DCGAN \citep{radford15}, and InfoGAN  \citep{chen16,signnet}. Following \cite{begus19}, we model language acquisition as learning of a dependency between the latent space and generated outputs. We introduce a network that forces the Generator to output data such that information is retrievable from its acoustic outputs and propose a new structure of the latent variables that allows featural learning and a very low-dimension vector representation of lexical items. Lexical learning emerges in an unsupervised manner from the architecture: the most efficient way for the Generator network to output acoustic data such that unique information is retrievable from its data is to encode unique information in its acoustic outputs such that latent codes coincide with lexical items in the training data. The result is thus a deep convolutional neural network that takes latent codes and variables  and outputs innovative data that resembles training data distributions as well as learns to associate lexical items with unique representations. 

Four experiments tested lexical learning in ciwGAN and fiwGAN architectures trained on tokens of five, ten, eight, and 6,229 sliced lexical items in raw audio format from a highly variable database --- TIMIT. The paper proposes that in smaller models lexical learning can be evaluated with multinomial logistic regression on generated data. Evidence of lexical learning is present in all four experiments. It appears that the Generator learns to associate lexical items with unique latent code --- categorical (as in ciwGAN) or featural (as in fiwGAN). By manipulating the values of latent codes to value 2, the networks output unique lexical items for each unique code and reach accuracy that ranges from  98\% to 26\% in the five-word model. To replicate the results and test learning on a higher number of lexical items, the paper presents evidence that the model learns to associate unique latent codes with lexical items in the 10-words model as well, with only one exception. When trained on the entire TIMIT, the Generator outputs one clearly identifiable lexical item in approximately 41\% of latent codes tested. The paper also proposes a technique for following how the network learns representations as training progresses. We can directly observe how the network transforms an output that violates training data into an output that conforms to it by keeping the latent space constant as training progresses. 

The fiwGAN architecture features, to our knowledge, a new proposal within the InfoGAN architecture: to model classification with featural codes instead of one-hot vectors. This shift yields  the potential to model featural learning and higher-level classification (i.e.~phonetic/phonological features and unique lexical representations) simultaneously. The paper presents evidence suggesting that the network might use some feature values to encode phonetic/phonological properties such as presence of [s]. Regression models suggest that $\phi_1$ is associated with presence of [s] in the output (and simple probabilistic calculation reveals about 8.6\% probability that the distribution is due to chance). 
The strongest evidence for simultaneous lexical and featural learning comes from innovative outputs in the fiwGAN architecture. The network trained on lexical items that lack a sequence of a fricative and a stop [st] altogether outputs an innovative sequence \emph{start} or \emph{sart}. These innovative outputs can be analyzed as adding a segment [s] (from \emph{suit}) to \emph{dark}, likely under the influence of the fact that $\phi_1$ represents presence of [s].

Innovative outputs that violate training data are informative for both computational models of language acquisition as well as for our understanding of what types of dependencies the networks are able to learn.   We discuss several cases of innovative outputs. Some innovations are motivated by training data distributions (e.g.~\emph{sear}) and reveal how the networks treat acoustically similar lexical items. For other innovative outputs, such as \emph{watery}, the training data contains no apparent motivations. We also track changes from innovative to conforming outputs as training progresses.

We argue that innovative outputs are linguistically interpretable and acoustically very similar to actual speech data that is absent from the training data. For example, an innovative [st] sequence in \emph{start} corresponds directly to human outputs with this sequence that were never part of the training data. Further  comparisons of this type should yield a better understanding on how the combinatorial principle in human language can arise without language-specific parameters in a model.

The paper also discusses how internal representations in deep convolutional networks can be identified and explored.
 We argue that by setting the latent values substantially beyond the training range (as suggested for phonological learning in \citealt{begus19}), the Generator almost exclusively outputs one unique lexical item per each unique featural code (with only one exception) in the fiwGAN architecture on eight lexical items. In other words, for very high values of the featural code ($\phi$), lexical learning appears to be near categorical.  The variability of the outputs is minimal at such high values (e.g.~15). It appears that setting the featural code to such extreme values reveals the underlying representation of each featural code. This property is highly desirable in a model of language acquisition and has the potential to reveal the underlying learned representations in the GAN architecture. 

Several further experiments and improvements to the model are left to future work. The proposed architecture allows testing of any property of the data and probing the learned representation behind each latent code (or variable; as per \citealt{begus19}). For example, we can train the network on $n$ latent codes or features, which effectively means we are forcing the network to learn $n$ (or $2^n$) most informative categories in the data. Manipulating the latent codes to marginal levels represents a technique to test which properties about the data the network learns in a generative fashion. We can thus test how deep convolutional networks learn informative properties about the data and directly observe what those categories are. Representation of any process in speech can thus be modeled  (for an identity-based pattern, see \citealt{begus20identity}). Second, the current paper only analyzes the Generator network in the generative fashion, but the architecture also allows an analysis of the  Q-network  in discriminative lexical learning tasks (such as ABX). An experiment with novel unobserved test data fed to the Q-network would reveal the model's ability to assign unique codes to individual lexical items. Such a network would be highly desirable in an unsupervised lexical learning setting or for   very low dimension acoustic word embedding tasks. Third, the paper proposes a technique to follow lexical learning as training progresses (see Section \ref{ciw5}). A superficial comparison between lexical learning in the proposed models and language-acquiring children could yield further insights into the computational modeling of language acquisition (for an overview of the literature on how lexical learning progresses in language acquisition, see \citealt{gaskell09}). The proposed architecture can also be used for testing lexical learning when trained on adult-directed vs.~child-directed speech corpora.  Finally, future directions should also include developing a model that could parse lexical items from a continuous acoustic speech stream and improving the model's overall performance.

The proposed model of lexical learning has several further implications. Dependencies in speech data are significantly better understood than dependencies in visual data. A long scientific tradition of studying dependencies in phonetic and phonological data in human languages yields an opportunity to use linguistic data to probe the types of dependencies deep neural networks can or cannot learn \citep{begus20identity}. The proposed architectures allow us to probe what types of dependencies the networks can learn, how they encode unique information in the latent space, and how  self-organization of retrievable information emerges in the GAN architecture. The models also have some basic implications for unsupervised text-to-speech generation tasks: manipulating the latent variables to specific values results in the Generator outputting desired lexical items.

\subsubsection*{Acknowledgements}
This research was funded by a grant to new faculty at the University of Washington and the University of California, Berkeley. I would like to thank  Sameer Arshad for slicing data from the TIMIT database and Ella Deaton for annotating data.

\subsubsection*{Declaration of interests}

The author declares no known competing financial interests or personal relationships that could have appeared to influence the work reported in this paper.

\newpage
  \bibliographystyle{elsarticle-harv} 
\bibliography{begusGANbib.bib}

\clearpage
\appendix
\section{Raw counts}

\begin{table}[H]\centering
\begin{tabular}{ccccccc}
\hline\hline
&else  & rag & oily &water & year & suit \\\hline
$c_1$&    7 &    0   &  0   &  0  &  21&    72 \\
$c_2$&  18   &  0    & 0     &0&    70&    12 \\
$c_3$&    6 &    0    &10   & 84    & 0    & 0 \\
$c_4$&   13  &   0  &  26  &  61    & 0  &   0 \\
$c_5$& 2    &98    & 0 &    0   &  0  &   0 \\
\hline\hline

\end{tabular}
\caption{Raw counts for outcomes of the ciwGAN model trained on five lexical items after 8,011  steps (Section \ref{8011}). The outcomes are coded as described in fn.~\ref{fntr}.}
\label{app1}
\end{table}

\begin{table}\centering
\begin{tabular}[H]{ccccccc}
\hline\hline
&else  & rag & oily &water & year & suit \\\hline
$c_1$&    17    & 0   &  0     &0     &9   & 74  \\
$c_2$&     13 &    0  &   0   &  0  &  71 &   16  \\
$c_3$&     10   &  0   & 12   & 78   &  0  &   0  \\
$c_4$&      15  &   0  &  37    &48 &    0    & 0  \\
$c_5$&     6   & 92   &  2    & 0 &    0 &    0  \\
\hline\hline

\end{tabular}
\caption{Raw counts for outcomes of the ciwGAN model trained on five items after 19,244  steps (Section \ref{19244}). The outcomes are coded as described in fn.~\ref{fntr}.}
\label{app2}
\end{table}

\begin{table}[H]\centering
\begin{tabular}{cccccccccccc}
\hline\hline

 &else  & suit&    rag  &  ask  &carry &  like &greasy   &year  & oily  &water  & dark \\\hline
c$_1$    &29&      0    & 21    &  0   &   1  &    0 &     0     & 0    &  0 &     0   &  49 \\
c$_2$&    31   &   0     & 0   &   0   &   0    &  4     & 0  &    0   &   8   &  57&      0 \\
c$_3$ &   37    &  0    &  0   &   0   &  10   &   1    &  0  &   11    & 39&      2     & 0 \\
c$_4$  &  20  &    0   &   8 &     0&      1    &  1     & 0    & 65   &   0 &     0      &5 \\
 c$_5$   & 1    &  0    &  0   &   0  &    0    &  0    & 99     & 0   &   0   &   0 &     0 \\
 c$_6$   &10  &    0  &    0   &  26   &   0    & 64 &     0 &     0 &     0 &     0    &  0 \\
c$_7$ &   20    &  0     & 0    &  0  &   76 &     0     & 0   &   4    &  0   &   0   &   0 \\
c$_8$ &   23  &    0 &     0    & 58   &   0  &   18    &  0    &  0 &     0  &    0&      1 \\
c$_9$ &   63    &  0  &    1   &   0&      1   &   2  &    0&      2&      7   &  18 &     6\\ 
c$_{10}$   &  8    & 92 &     0      &0     & 0  &    0    &  0   &   0    &  0 &     0 &     0 \\
\hline\hline  

\end{tabular}
\caption{Raw counts for outcomes of the ciwGAN model trained on ten items after 27,103  steps (Section \ref{ciw10}). The outcomes are coded as described in fn.~\ref{code10}.}
\label{app3}
\end{table}

\begin{table}[H]\centering
\begin{tabular}{cccccccccc}
\hline\hline

&  else  & dark   & ask &  suit& greasy   &year&  water & carry&   like \\\hline
 $[0, 0, 0]$   &43      &0   &  15   &   0   &  17  &    3&      3     & 2   &  17 \\
  $[0, 0, 2]$  &18&      1 &     0  &   23   &   0    & 11  &    0&     47 &     0 \\
$[0, 2, 0]$&   32  &    0  &    0   &   0      &0  &    0 &    32  &    0&     36\\ 
$[0, 2, 2]$&   18   &   0   &   0&      0   &   0 &    50    & 32   &   0     & 0 \\
 $[2, 0, 0]$   &20    &  1 &    19     & 0&     47   &  13    &  0    &  0  &    0 \\
 $[2, 0, 2]$  & 20     & 7     & 0     &46&      0  &    0 &     0  &   27  &    0\\
  $[2, 2, 0]$ & 15  &    1 &    61 &     0    & 22&      0&      0&      1   &   0\\ 
 $[2, 2, 2]$ &  25   &  54&      0     &19 &     0  &    0    &  0 &     2     & 0 \\
\hline\hline  

\end{tabular}
\caption{Raw counts for outcomes of the fiwGAN model trained on eight items after 20,026  steps (Section \ref{fiwGANon8}). The outcomes are coded as described in fn.~\ref{code8}.}
\label{app4}
\end{table}

\end{document}